\RequirePackage{etoolbox}
\csdef{input@path}{%
 {sty/}
 {img/}
}%
\csgdef{bibdir}{bib/}

\documentclass[ba]{imsart}
\pubyear{0000}
\volume{00}
\issue{0}
\doi{0000}
\firstpage{1}
\lastpage{1}

\usepackage{xr-hyper}
\usepackage{amsthm}
\usepackage{amsmath}
\usepackage{amssymb,amsfonts}
\usepackage{natbib}
\usepackage{graphicx}
\usepackage[english]{babel}
\usepackage{mathtools} 
\usepackage{algorithm,algorithmic}
\usepackage{bbm}
\usepackage{caption}
\usepackage{subfig}
\usepackage{enumitem}
\usepackage{booktabs}
\usepackage{xcolor}

\usepackage[colorlinks,citecolor=blue,urlcolor=blue,filecolor=blue,backref=page]{hyperref}

\startlocaldefs

\newcommand{\dataObs}{x_0}
\newcommand{\lik}{L}
\newcommand{\likHat}{\hat{\lik}}
\newcommand{\pmodel}{p}
\newcommand{\pmodelHat}{\hat{\pmodel}}
\newcommand{\pmarg}{p}
\newcommand{\pmargHat}{\hat{p}}
\newcommand{\ppost}{p}
\newcommand{\prior}{p}
\newcommand{\postHat}{\hat{p}}
\newcommand{\xtheta}{x^\theta}
\newcommand{\Xtheta}{X^\theta}
\newcommand{\xm}{x^m}
\newcommand{\Xm}{X^m}
\newcommand{\ntheta}{n_\theta}

\newcommand{\rHat}{\hat{r}}
\newcommand{\hHat}{\hat{h}}
\newcommand{\coeffHat}{\hat{\beta}}
\newcommand{\regCoeffHat}{\hat{\beta}_{\small \mathrm{reg}}}
\newcommand{\J}{\mathcal{J}}
\newcommand{\Jbar}{\tilde{\J}}
\newcommand{\sKL}{\mathrm{sKL}}
\newcommand{\DeltaKL}{\Delta_{\small \mathrm{sKL}}}

\newcommand{\lambdaMin}{\lambda_{\small \mathrm{min}}}
\newcommand{\risk}{\mathcal{R}}

\newcommand{\pdata}{p_x}
\newcommand{\pnoise}{p_y}
\newcommand{\relerr}{\mathcal{RE}}
\newcommand{\relerrsl}{\mathcal{RE}_{\small \mathrm{SL}}}
\newcommand{\Deltarelerr}{\Delta_{\small \mathrm{rel-error}}}

\newcommand{\ud}{\,\mathrm{d}}

\def\P{\mathbb{P}}

\newcommand{\E}{\mathbb{E}}

\DeclareMathOperator*{\argmin}{arg\,min}
\DeclareMathAlphabet{\mathpzc}{OT1}{pzc}{m}{it}

\newcommand{\smref}{Supplementary Material~}

\endlocaldefs

\newif\ifsepsuppl
\sepsupplfalse 

\ifsepsuppl 
\fi

\begin{document}


\begin{frontmatter}
\title{Likelihood-free inference by ratio estimation}

\runtitle{LFIRE}

\begin{aug}
\author{\fnms{Owen} \snm{Thomas}\thanksref{addr1}\ead[label=e1]{o.m.t.thomas@medisin.uio.no}},
\author{\fnms{Ritabrata} \snm{Dutta}\thanksref{addr2}\ead[label=e2]{ritabrata.dutta@warwick.ac.uk}},
\author{\fnms{Jukka} \snm{Corander}\thanksref{addr1}\ead[label=e3]{jukka.corander@medisin.uio.no}},
\author{\fnms{Samuel} \snm{Kaski}\thanksref{addr4}\ead[label=e4]{samuel.kaski@aalto.fi}}
\and
\author{\fnms{Michael U.} \snm{Gutmann}\thanksref{addr5,t1}\ead[label=e5]{michael.gutmann@ed.ac.uk}}

\runauthor{O. Thomas et al.}

\address[addr1]{ Department of Biostatistics, University of Oslo, Norway.
\printead{e1}\\ \printead{e3}}  
\address[addr2]{ Department of Statistics, University of Warwick, UK.
\printead{e2}}

\address[addr4]{ Helsinki Institute for Information Technology, Department of Computer Science, Aalto University, Finland.
\printead{e4}}
\address[addr5]{School of Informatics, University of Edinburgh, UK.
    \printead{e5}
}
\thankstext{t1}{(Corresponding author)}

\end{aug}

\begin{abstract}
We consider the problem of parametric statistical inference when
likelihood computations are prohibitively expensive but sampling from
the model is possible. Several so-called likelihood-free methods have
been developed to perform inference in the absence of a likelihood
function. The popular synthetic likelihood approach infers the
parameters by modelling summary statistics of the data by a Gaussian
probability distribution. In another popular approach called
approximate Bayesian computation, the inference is performed by
identifying parameter values for which the summary statistics of the
simulated data are close to those of the observed data. Synthetic
likelihood is easier to use as no measure of ``closeness'' is required
but the Gaussianity assumption is often limiting. Moreover, both
approaches require judiciously chosen summary statistics. We here
present an alternative inference approach that is as easy to use as
synthetic likelihood but not as restricted in its assumptions, and
that, in a natural way, enables automatic selection of relevant
summary statistic from a large set of candidates. The basic idea is to
frame the problem of estimating the posterior as a problem of
estimating the ratio between the data generating distribution and the
marginal distribution. This problem can be solved by logistic
regression, and including regularising penalty terms enables automatic
selection of the summary statistics relevant to the inference task. We
illustrate the general theory on canonical examples and employ it to
perform inference for challenging stochastic nonlinear dynamical
systems and high-dimensional summary statistics.
\end{abstract}


\end{frontmatter}


\section{Introduction}
\label{sec:intro}
We consider the problem of estimating the posterior probability
density function (pdf) of some model parameters $\theta \in
\mathbb{R}^d$ given observed data $\dataObs \in \mathcal{X}$ when
computation of the likelihood function is too costly but data can be
sampled from the model. In particular, we assume that the model
specifies the data generating pdf $\pmodel(x|\theta)$ not explicitly,
e.g. in closed form, but only implicitly in terms of a stochastic
simulator that generates samples $x$ from the model
$\pmodel(x|\theta)$ for any value of the parameter $\theta$. The
simulator can be arbitrarily complex so that we do not impose any
particular conditions on the data space $\mathcal{X}$. Such
simulator-based (generative) models are used in a wide range of
scientific disciplines to simulate different aspects of nature on the
computer, for example in genetics \citep{Pritchard1999, Arnold2018},
ecology \citep{Wood_2010, Siren2018}, or epidemiology of infectious
diseases \citep{Tanaka2006, Corander2017}.

Denoting the prior pdf of the parameters by $\prior(\theta)$, the
posterior pdf $\ppost(\theta | \dataObs)$ can be obtained from Bayes'
formula,
\begin{align}
  \ppost(\theta|x) &= \frac{\prior(\theta) \pmodel(x | \theta)}{p(x)}, & p(x) &= \int \prior(\theta) \pmodel(x | \theta) \ud \theta,
\label{eq:Bayes}
\end{align}
for $x=\dataObs$. Exact computation of the posterior pdf is, however,
impossible if the likelihood function $L(\theta) \propto
\pmodel(\dataObs | \theta)$ is too costly to compute. Several
approximate inference methods have appeared for simulator-based
models. They are collectively known as likelihood-free inference
methods, and include approximate Bayesian computation
\citep{Tavare1997, Pritchard1999, Beaumont2002} and the synthetic
likelihood approach \citep{Wood_2010}. For a comprehensive
introduction to the field, we refer the reader to the review papers by
\citet{Beaumont2010, Hartig2011, Marin_2012, Lintusaari_2017,
  Sisson2018}.

Approximate Bayesian computation (ABC) relies on finding parameter
values for which the simulator produces data that are similar to the
observed data. Similarity is typically assessed by reducing the
simulated and observed data to summary statistics and comparing their
distance. While the summary statistics are classically determined by
expert knowledge about the problem at hand, there have been recent
pursuits in choosing them in an automated manner
\citep{Aeschbacher_2012, Fearnhead_2012, Blum_2013, Gutmann_2014,
  Gutmann2018}. While ABC can be considered to implicitly construct a
nonparametric approximation of $\pmodel(x|\theta)$
\citep[e.g.][]{Hartig2011, Lintusaari_2017}, a wide range of
parametric surrogate models are being used to accelerate
the inference or improve its accuracy. The models employed include
regression models and neural networks, Gaussian processes as well as
normalising flows \citep{Beaumont2002, Blum2010b, Wilkinson_2014,
  Gutmann_2016, Papamakarios_2016, papamakarios2017masked,
  papamakarios2019, Chen2019}. Synthetic likelihood, on the other
hand, assumes that the summary statistics for a given parameter value
follow a Gaussian distribution \citep{Wood_2010}. The synthetic
likelihood approach is applicable to a diverse set of problems
\citep{Meeds_2014,Price2017}, but the Gaussianity assumption may not
always hold and the original method does not include a mechanism for
choosing summary statistics automatically.

In this paper, we propose (1) a framework, ``LFIRE'', and (2) a
practical method, ``linear LFIRE'', to directly approximate the
posterior distribution in the absence of a tractable likelihood
function.\footnote{The ideas in this paper were first communicated on
  arXiv in 2016 \citep{Thomas2016}. The reader may wonder about the
  several years difference between the arXiv paper and this
  paper. This is largely due to three review periods that took 8, 9,
  and 7 months, respectively, and the introduction of a new first
  author. The core content has stayed the same. We thus would like to
  ask you to please also acknowledge \citep{Thomas2016} when citing
  this paper.}  As we will see, the proposed approach includes
the synthetic likelihood as a special case and further enables
automatic selection of summary statistics in a natural way.

The basic idea is to frame the original problem of estimating the
posterior as a problem of estimating the ratio $r(x,\theta)$ between
the data generating pdf $\pmodel(x | \theta)$ and the marginal
distribution $p(x)$, in the context of a Bayesian belief update
\begin{equation}
  r(x,\theta) =  \frac{\pmodel(x | \theta)}{\pmarg(x)}.
\label{eq:rdef}
\end{equation}
By definition of the posterior distribution, an estimate $\rHat(x,\theta)$ for the ratio
implies an estimate $\postHat(\theta|\dataObs)$ for the posterior,
\begin{equation}
  \postHat(\theta|\dataObs) = \prior(\theta) \rHat(\dataObs,\theta).
  \label{eq:postHat}
\end{equation}
In addition, the estimated ratio also yields an estimate
$\likHat(\theta)$ of the likelihood function,
\begin{equation}
  \likHat(\theta) \propto \rHat(\dataObs,\theta),
  \label{eq:Lhat}
\end{equation}
as the denominator $p(x)$ in the ratio does not depend on $\theta$. We
can thus perform likelihood-free inference by ratio estimation, and we
call this \emph{framework} in short ``LFIRE''.

In the LFIRE framework, other distributions than the marginal $p(x)$ can also be used in the
denominator, in particular if approximating the likelihood function or
identifying its maximiser is the goal. While we do not further address
the question of what distributions can be chosen for estimation of the
posterior, initially it seems reasonable to prefer distributions
that have heavier tails than $p(x|\theta)$ in the numerator because of
stability reasons.

Closely related work was done by \citet{Pham_2014} and
\citet{Cranmer_2015} who estimated likelihood
ratios. \citet{Pham_2014} estimated the ratio between the likelihoods
of two parameters appearing in the acceptance probability of the
Metropolis-Hastings MCMC sampling scheme. If we used the approximate
posterior distribution in Equation \eqref{eq:postHat} to estimate the
acceptance probability, we would also end up with a density ratio that
can be used for MCMC sampling. A key difference is that our approach
results in estimates of the posterior and not in a single accepted, or
rejected, parameter value. \citet{Cranmer_2015} estimated the ratio
between the likelihood at a freely varying parameter value and a fixed
reference value in the context of frequentist inference. The goals are
thus somewhat different, which, as we will see, account well for the
differences in the results in our empirical comparison in Section
\ref{sec:carl_comparison}.

Since we have first communicated the LFIRE framework as an arXiv paper
\citep{Thomas2016}, there have been a number of developments within
this framework. For instance, \citet{Dinev2018} tailored the approach
to the special case of time-series models, \citet{RogerSmith2018}
adapted the framework to a sequential population Monte Carlo scheme
with adaptive proposals, and \citet{Hermans2020} perform (sequential)
LFIRE for amortised likelihood-free MCMC sampling. \citet{Durkan2020}
discuss how inference methods in the LFIRE framework relate to
conditional density estimation methods, in particular drawing
connections between the work by \citet{Hermans2020} and
\citet{Greenberg2019}.
  
There are several methods in the literature available for the
estimation of density ratios \citep[e.g.][]{Gutmann2011b,
  Sugiyama_2012, Izbicki_2014}, of which estimation through logistic
regression is widely used and has some favourable asymptotic
properties
\citep{Geyer_1994,Qin_1998,Cheng_2004,Bickel_2007}. Logistic
regression is very closely related to probabilistic classification and
we use it in the paper to estimate the ratio $r(x,\theta)$. 

Logistic regression and probabilistic classification have been
employed before to address other computational problems in
statistics. \citet{Gutmann_2012} used this kind of ``contrastive
learning'' to estimate unnormalised models and \citet{Goodfellow_2014}
employed it for training neural networks to generate samples similar
to given reference data. More general methods for ratio estimation are
also used for training such neural networks \citep[see e.g.\ the
  review by][]{Mohamed2016}, and they were used before to estimate
unnormalised models \citep{Pihlaja2010,Gutmann2011b}. Classification
has been shown to yield a natural distance function in terms of the
classifiability between simulated and observed data, which can be used
for ABC \citep{Gutmann_2014, Gutmann2018}. While this earlier approach
is very general, the classification problem is difficult to set up
when the observed data consist of very few data points only. The
related work by \citet{Pham_2014} and \citet{Cranmer_2015} and the
method proposed in this paper do not have this shortcoming.

The rest of the paper is organised as follows: Section
\ref{sec:classification} presents the details on how to generally
estimate the ratio $r(x,\theta)$ and hence the posterior by logistic
regression. In Section \ref{sec:model}, we model the ratio as a linear
superposition of summary statistics, yielding the ``linear LFIRE''
method, and show that this assumption corresponds to an exponential family
approximation of the intractable model pdf. As Gaussian distributions
are part of the exponential family, our approach thus includes the
synthetic likelihood approach as a special case. We then show in
Section \ref{sec:selection} that including a penalty term in the
logistic regression enables automatic selection of relevant summary
statistics. In Section \ref{sec:validation}, we validate the resulting
method on canonical examples, and in Sections \ref{sec:application}
and \ref{sec:highdimsummstats}, we apply it to challenging inference
problems in ecology, weather forecasting, and cell proliferation
modelling. All simulation studies include a comparison with the
synthetic likelihood approach, with their relative computational costs
analysed in Section \ref{sec:timing}. We find that the new method
yielded consistently more accurate inference results than synthetic
likelihood.

\section{Posterior estimation by logistic regression}
\label{sec:classification}
\begin{figure}
\includegraphics[scale=0.8]{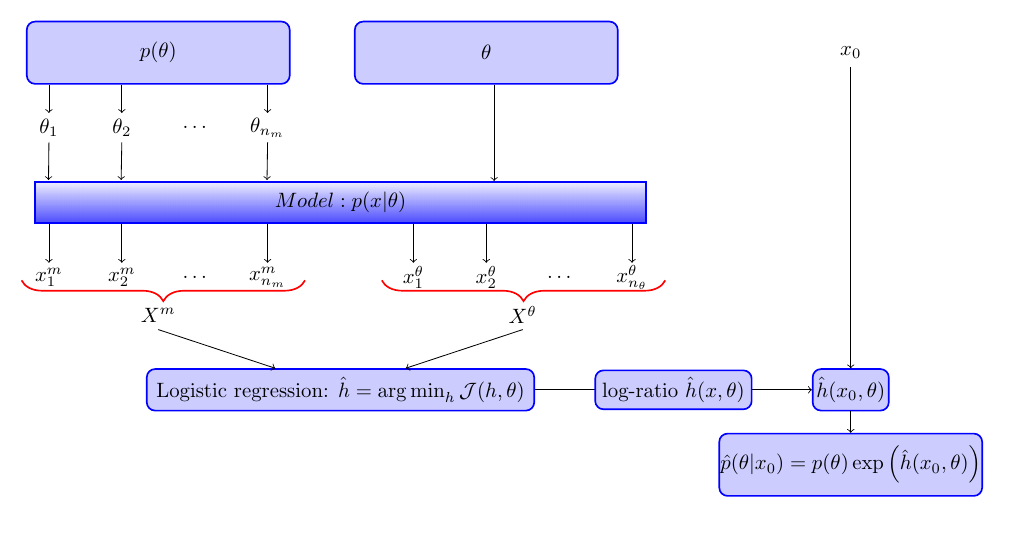}
\caption{A schematic view of likelihood-free inference by ratio
  estimation (LFIRE) by means of logistic regression, as explained in
  Equations \eqref{eq:marginal_sampling} to
  \eqref{eq:nonparametricEstimates}.}
\label{fig:diagram}
\end{figure}

We here show that the ratio $r(x,\theta)$ in Equation \eqref{eq:rdef}
can be estimated by logistic regression, which yields estimates for
the posterior and the likelihood function together with Equations
\eqref{eq:postHat} and \eqref{eq:Lhat}. Figure \ref{fig:diagram}
provides an overview.

As we assumed working with a simulator-based model, we can generate
data from the pdf $\pmodel(x|\theta)$ in the numerator of the ratio
$r(x,\theta)$; let $\Xtheta = \{\xtheta_i\}_{i=1}^{\ntheta}$ be such a
set with $\ntheta$ independent samples generated with a fixed value of
$\theta$. Additionally we can also generate data from the marginal pdf
$\pmarg(x)$ in the denominator of the ratio; let $\Xm =
\{\xm_i\}_{i=1}^{n_{m}}$ be such a set with $n_m$ independent
samples. As the marginal $p(x)$ is obtained by integrating out
$\theta$, see Equation \eqref{eq:Bayes}, the samples can be obtained
by first sampling from the joint distribution of $(x,\theta)$ and then
ignoring the sampled parameters,
\begin{align}
  \label{eq:marginal_sampling}
 \theta_i &\sim \prior(\theta), &  x^m_i &\sim \pmodel(x|\theta_i).
\end{align}

We now formulate a classification problem where we aim to
determine whether some data $x$ were sampled from $\pmodel(x|\theta)$
or from $p(x)$. This classification problem can be solved via
(nonlinear) logistic regression \citep[e.g.][]{Hastie_2001}, where
the probability for $x$ to belong to $\Xtheta$, for instance, is
parametrised by some nonlinear function $h(x)$,
\begin{align}
  \P( x \in \Xtheta; h ) &= \frac{1}{1+\nu \exp(-h(x))}, 
\end{align}
with $\nu= n_m/ \ntheta$ compensating for unequal class sizes. A
larger value of $h$ at $x$ indicates a larger probability for $x$ to
originate from $\Xtheta$. A suitable function $h$ is typically found
by minimising the loss function $\J$ on the training data $\Xtheta$
and $\Xm$,
\begin{equation}
  \J(h,\theta) = \frac{1}{\ntheta+n_m} \left\{ \sum_{i=1}^{\ntheta}
    \log\left[1+\nu \exp(-h(\xtheta_i))\right] + \sum_{i=1}^{n_m}
    \log\left[1+\frac{1}{\nu} \exp(h(\xm_i))\right] \right\}.
\label{eq:Jthetah}
\end{equation}
The dependency of the loss function on $\theta$ is due to the
dependency of the training data $\Xtheta$ on $\theta$. 

We prove in \smref \ref{proof:logr-opt} that for large $n_m$ and $\ntheta$, the
minimising function $h^\ast$ is given by the log-ratio between
$\pmodel(x|\theta)$ and $\pmarg(x)$, that is
\begin{equation}
  h^\ast(x,\theta) = \log r(x,\theta).
\label{eq:hoptimal}
\end{equation}
The proof shows that this holds for all $\theta$ for which we provide
data in the classification. This means that we can average the loss
function with respect to some distribution $f(\theta)$ defined on a
domain where we would like to evaluate the ratio, which corresponds to
performing the classification in the joint $x, \theta$ space. Choosing
for $f(\theta)$ the prior $p(\theta)$, we would classify between data
from $p(x|\theta)p(\theta)$ and $p(x)p(\theta)$, as recently done by
\citet{Hermans2020}, which enables amortisation of the computations
with respect to $\theta$.

For finite sample sizes $n_m$ and $\ntheta$, the minimising function $\hHat$,
\begin{equation}
  \hHat = \argmin_{h} \J(h,\theta),
\label{eq:hthetaHat}
\end{equation}
thus provides an estimate $\rHat(x,\theta)$ of the ratio $r(x,\theta)$,
\begin{equation}
\rHat(x,\theta) = \exp(\hHat(x,\theta)),
\label{eq:rHat}
\end{equation}
and Equations \eqref{eq:postHat} and \eqref{eq:Lhat} yield the
corresponding estimates for the posterior and likelihood function,
respectively,
\begin{align}
  \postHat(\theta|\dataObs)&=
  \prior(\theta)\exp(\hHat(\dataObs,\theta)),& \likHat(\theta) \propto
  \exp(\hHat(\dataObs,\theta)).
\label{eq:nonparametricEstimates}
\end{align}
In case samples from the posterior are needed, we can use standard
sampling schemes with $\postHat(\theta|\dataObs)$ as the target pdf
\citep{Andrieu_2009}, for instance MCMC \citep{Hermans2020}. The
estimates can also be used together with Bayesian optimisation
\citep{Gutmann_2016} or history matching \citep{Wilkinson_2014} to
accelerate the inference.
When estimating the posterior or likelihood function as outlined
above, the sample sizes $n_m$ and $\ntheta$ are entirely under our
control. Their values reflect the trade-off between computational and
statistical efficiency. We note that both $\Xtheta$ and $\Xm$ can be
constructed in a perfectly parallel manner. Moreover, while $\Xtheta$
needs to be constructed for each value of $\theta$, $\Xm$ is
independent of $\theta$ and needs to be generated only once.

Different models can be used for probabilistic classification;
equivalently, different assumptions can be made on the family of
functions to which the log-ratio $h$ belongs. While non-parametric
families or deep architectures can be used \citep{Dinev2018}, we next
consider a simple parametric family that is spanned by a set of
summary statistics, yielding a particular inference method of the
more general LFIRE framework.
\section{Exponential family approximation}
\label{sec:model}
We here restrict the search in Equation \eqref{eq:hthetaHat} to
functions $h$ that are members of the family spanned by $b$ summary
statistics $\psi_i(x)$, each mapping data $x \in \mathcal{X}$ to $\mathbb{R}$,
\begin{equation}
  h(x) = \sum_{i=1}^b  \beta_i \psi_i(x) = \beta^\top \psi(x),
\label{eq:span}
\end{equation}
with $\beta_i \in \mathbb{R}$, $\beta = (\beta_1,\ldots, \beta_b)$,
and $\psi(x) = (\psi_1(x), \ldots, \psi_b(x))$. This corresponds to
performing logistic regression with a linear basis expansion
\citep{Hastie_2001}. The observed data $\dataObs$ may be used in the
definition of the summary statistics, as for example with the Ricker
model in Section \ref{sec:application}, and thus influence the
logistic regression part of the likelihood-free inference pipeline in
Figure \ref{fig:diagram} (not shown in the figure). Given the linear
nature of the model in \eqref{eq:span}, we may call this instance of
the LFIRE inference principle ``linear LFIRE''.

When we assume that $h(x)$ takes the functional form in Equation
\eqref{eq:span}, estimation of the ratio $r(x,\theta)$ boils down
to the estimation of the coefficients $\beta_i$. This is done by
minimising $J(\beta,\theta) = \J(\beta^\top\psi,\theta)$ with respect to $
\beta$,
\begin{align}
  \coeffHat(\theta) &=\argmin_{\beta \in \mathbb{R}^b} J(\beta,\theta),\\
 J(\beta,\theta) &=\frac{1}{\ntheta+n_m}\left\{\sum_{i=1}^{\ntheta} \log\left[1+\nu \exp(- \beta^\top \psi_i^\theta  )\right] + \sum_{i=1}^{n_m} \log\left[1+\frac{1}{\nu} \exp(\beta^\top \psi_i^m)\right]\right\}
\label{eq:min_negloglikelihood}
\end{align}
The terms $\psi_i^\theta = \psi(\xtheta_i)$ and $\psi_i^m =
\psi(\xm_i)$ denote the summary statistics of the simulated data sets
$\xtheta_i \in \Xtheta$ and $\xm_i \in \Xm$, respectively. The
estimated coefficients $\coeffHat$ depend on $\theta$ because the training data
$\xtheta_i \in \Xtheta$ depend on $\theta$. With the model assumption
in Equation \eqref{eq:span}, the estimate for the ratio in Equation
\eqref{eq:rHat} thus becomes
\begin{align}
  \rHat(x,\theta) &= \exp(\coeffHat(\theta)^\top\psi(x))
  \label{eq:rHatParametric}
\end{align}
and the estimates for the posterior and likelihood function in
Equation \eqref{eq:nonparametricEstimates} are
\begin{align}
  \postHat(\theta|\dataObs)&=
  \prior(\theta)\exp( \coeffHat(\theta)^\top\psi(\dataObs)),& \likHat(\theta) \propto
  \exp( \coeffHat(\theta)^\top\psi(\dataObs)),
  \label{eq:parametricEstimates}
\end{align}
respectively.

As $r(x,\theta)$ is the ratio between
$\pmodel(x|\theta)$ and $\pmarg(x)$, we can consider the estimate
$\rHat(x,\theta)$ in Equation \eqref{eq:rHatParametric} to provide an
implicit estimate $\pmodelHat(x|\theta)$ of the intractable model pdf
$\pmodel(x|\theta)$,
\begin{align}
  p(x|\theta) &\approx \pmodelHat(x|\theta), &
  \pmodelHat(x|\theta) &=\pmargHat(x) \exp(\coeffHat(\theta)^\top\psi(x)).
\label{eq:expFamilyApprox}
\end{align}
The estimate is implicit because we have not explicitly estimated the
marginal pdf $\pmarg(x)$. Importantly, the equation shows that
$\pmodelHat(x|\theta)$ belongs to the exponential family with
$\psi(x)$ being the sufficient statistics for the family, and
$\coeffHat(\theta)$ the vector of natural parameters.

In previous work, \citet{Wood_2010} in the synthetic likelihood
approach, as well as \citet{Leuenberger_2010}, approximated the model
pdf by a member from the Gaussian family. As the Gaussian family
belongs to the exponential family, the approximation in Equation
\eqref{eq:expFamilyApprox} includes this previous work as a special
case. Specifically, a synthetic likelihood approximation with summary
statistics $\phi$ corresponds to an exponential family approximation
where the summary statistics $\psi$ are the individual $\phi_k$, all
pairwise combinations $\phi_{k} \phi_{k'}$, $k \ge k'$, and a
constant. While in the synthetic likelihood approach, the weights of
the summary statistics are determined by the mean and covariance
matrix of $\phi$, in our approach, they are determined by the solution
of the optimisation problem in \eqref{eq:min_negloglikelihood}. Hence,
even if equivalent summary statistics are used, the two approaches can
yield different approximations if the summary statistics are actually
not Gaussian. We will see that for equivalent summary statistics,
relaxing the Gaussianity assumption typically leads to better
inference results.

\section{Data-driven selection of summary statistics}
\label{sec:selection}
The estimated coefficients $\coeffHat(\theta)$ are weights that
determine to which extent a summary statistic $\psi_i(x)$ contributes
to the approximation of the posterior. As the number of simulated data
sets $n_m$ and $\ntheta$ increases, the error in the estimates
$\coeffHat(\theta)$ decreases and the importance of each summary
statistic can be determined more accurately. Increasing the number of
simulated data sets, however, increases the computational cost too. As
an alternative to increasing the number of simulated data sets, we
here use an additional penalty term in the logistic regression to
determine the importance of each summary statistic.

This approach enables us to work with a large list of candidate
summary statistics and automatically select the relevant ones in a
data-driven manner. This makes the posterior inference more robust and
less dependent on subjective user input. Moreover, the selection of
summary statistics through regularisation can substantially increase
the interpretability of the inference: the number of data summaries
identified as relevant may be small enough to be examined by
statisticians and model experts individually, providing evidence-based
insight into which summary statistics are relevant to the scientific
question.

\begin{algorithm}[ht!]
\caption{Linear LFIRE by penalised logistic regression}
\begin{algorithmic}[1]
\label{algo:approx_posterior}
  \scriptsize
\STATE Consider b-dimensional summary statistics $\psi : x \in \mathcal{R} \mapsto \mathbb{R}^b$. 
\STATE Simulate $n_m$ samples $\{x_i^m\}_{i=1}^{n_m}$ from the marginal density $\pmarg(x)$.
\STATE To estimate the posterior pdf at parameter value $\theta$ do: 
\begin{enumerate}[label=\alph*.]
 \item Simulate $\ntheta$ samples
   $\{x_i^{\theta}\}_{i=1}^{n_{\theta}}$ from the model pdf
   $\pmodel(x|\theta)$
\item Estimate $\regCoeffHat(\theta,\lambda)$ by solving the
  optimisation problem in Equation \eqref{eq:lasso} for $\lambda \in
  [10^{-4}\lambda_0,\lambda_0]$ where $\lambda_0$ is the smallest
  $\lambda$ value for which $\regCoeffHat = 0$. 
\item Find the minimiser $\lambdaMin$ of the prediction risk
  $\risk(\lambda)$ in Equation \eqref{eq:predictionRisk} as
  estimated by ten-fold cross-validation, and set $\coeffHat(\theta) =
  \regCoeffHat(\theta,\lambdaMin)$.
\item Compute the value of the estimated posterior pdf $\postHat(\theta| \dataObs)$ according to Equation \eqref{eq:parametricEstimates}.
\end{enumerate}
For the results in this paper, we always used $\ntheta=n_m$. To implement steps b and c we used the R package `glmnet' \citep{Friedman_2009}.
\end{algorithmic}
\end{algorithm}

While many choices are possible, we use the $L_1$ norm of
the coefficients as penalty term, like in lasso regression
\citep{Tibshirani_1994}. The coefficients $\beta$ in the basis
expansion in Equation \eqref{eq:span} are thus determined as the solution
of a $L_1$-regularised logistic regression problem,
\begin{align}
  \regCoeffHat(\theta,\lambda) &=\argmin_{\beta \in \mathbb{R}^b} J(\beta,\theta) + \lambda \sum_{i=1}^b |\beta_i|.
\label{eq:lasso}
\end{align}
The value of $\lambda$ determines the degree of the
regularisation. Sufficiently large values cause some of the
coefficients to be exactly zero. Different schemes to choose $\lambda$
have been proposed that aim at minimising the prediction risk
\citep{Zou_2007, Wang_2007, Tibshirani_2012, Dutta_2012}. Following
common practice and recommendations \citep[e.g.\ ][]{Tibshirani_1994,
  Hastie_2001}, we here choose
$\lambda$ by minimising the prediction risk $\mathcal{R}(\lambda)$,
\begin{equation}
\mathcal{R}(\lambda) = \frac{1}{\ntheta+n_m}\left\{\sum_{i=1}^{\ntheta} \mathbbm{1}_{\Pi_{\lambda}(x_i^{\theta})<0.5} + \sum_{i=1}^{n_m} \mathbbm{1}_{\Pi_{\lambda}(x_i^{m})>0.5}\right\},
\label{eq:predictionRisk}
\end{equation}
estimated by ten-fold cross-validation, where $\Pi_{\lambda}(x) = \P( x \in \Xtheta; h(x) = \regCoeffHat(\theta,\lambda)^\top \psi(x))$. The minimising value
 $\lambdaMin$ determines the coefficient $\coeffHat(\theta)$,
\begin{equation}
  \coeffHat(\theta) = \regCoeffHat(\theta,\lambdaMin),
\end{equation}
which is used in the estimate of the density ratio in Equation
\eqref{eq:rHatParametric}, and thus the posterior and likelihood in
Equation \eqref{eq:expFamilyApprox}.
Algorithm~\ref{algo:approx_posterior} presents pseudo-code that
summarises the linear LFIRE procedure for joint summary statistics selection and
posterior estimation. Algorithm~\ref{algo:approx_posterior} is a special case of the scheme
described in Figure~\ref{fig:diagram} when $h(x)$ is a linear
combination of the summary statistics $\psi(x)$ as described in
Equation~\eqref{eq:span}.

The cross-validation adds computational cost and the
dependency of $\lambdaMin$ on $\theta$ can make more detailed
theoretical investigations more difficult. In order to reduce the cost
or to facilitate theoretical analyses, working with a fixed value
of $\lambda$ as, for example, \citet{An2019} for synthetic likelihood
with the graphical lasso may be appropriate.

\section{Validation on canonical low-dimensional problems}
\label{sec:validation}
We here validate and illustrate the presented theory on a set of
canonical inference problems widely considered in the likelihood-free
inference literature and empirically compare the proposed approach to an approach based on likelihood ratios.

\subsection{Gaussian distribution}
We illustrate the proposed inference method on the simple example of
estimating the posterior pdf of the mean of a Gaussian
distribution with known variance. The observed data $\dataObs$ is a single observation
that was sampled from a univariate Gaussian with mean $\mu_o = 2.3$
and standard deviation $\sigma_o=3$. Assuming a uniform prior
$\mathcal{U}(-20,20)$ on the unknown mean $\mu$, the log posterior
density of $\mu$ given $\dataObs$ is
\begin{align}\label{eq:Gaussian}
  \log{p(\mu|\dataObs)} &= \alpha_0(\mu)+ \alpha_1(\mu)\dataObs + \alpha_2(\mu)\dataObs^2
\end{align}
if $\mu \in (-20,20)$, and zero otherwise. The model is thus within the
family of models specified in Equation \eqref{eq:parametricEstimates}.
Coefficient $\alpha_0(\mu)$ equals
\begin{align}
\alpha_0(\mu) &= -\frac{\mu^2}{2\sigma_0^2}-\log{(\sqrt{2\pi \sigma_0^2})}-\log\left(\Phi\left(\frac{20-x_0}{\sigma_0}\right)-\Phi\left(\frac{-20-x_0}{\sigma_0}\right)\right),
\end{align}
where $\Phi$ is the cumulative distribution function of the standard
normal distribution, and the coefficients $\alpha_1(\mu)$ and
$\alpha_2(\mu)$ are
\begin{align}
\alpha_1(\mu) &= \frac{\mu}{\sigma_0^2},& \alpha_2(\mu) &= -\frac{1}{2\sigma_0^2}.
\end{align}

\begin{figure}
  \centering \subfloat[]{\includegraphics[width=0.45\textwidth]{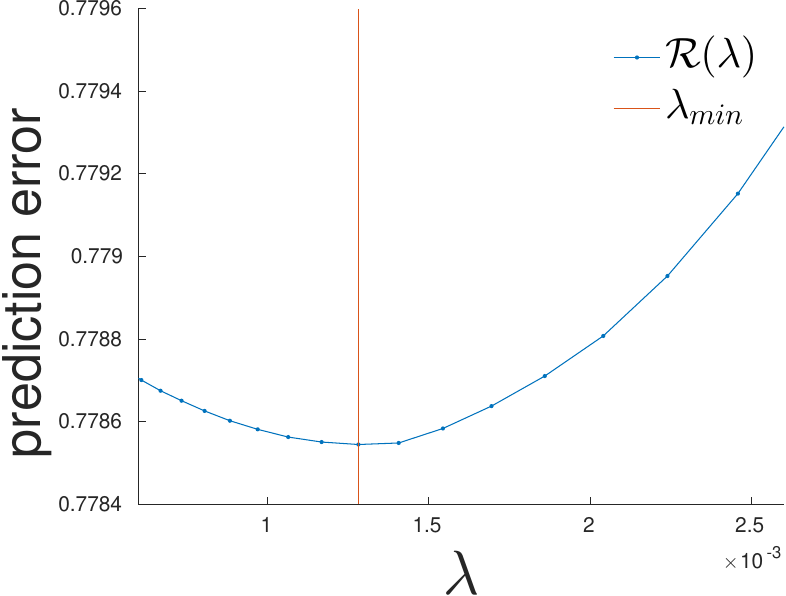}\label{fig:f1a}}
  \\ \subfloat[]{\includegraphics[width=0.45\textwidth]{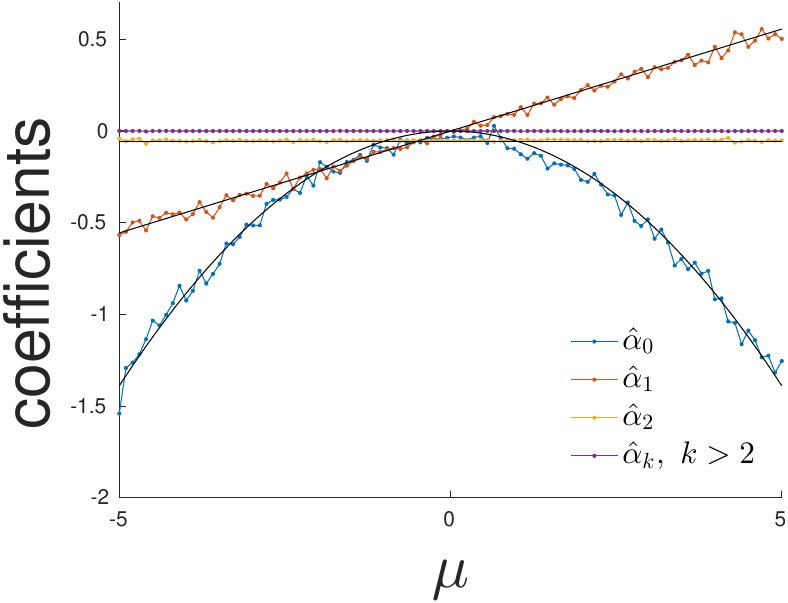}\label{fig:f1b}}
  \hfill \subfloat[]{\includegraphics[width=0.45\textwidth]{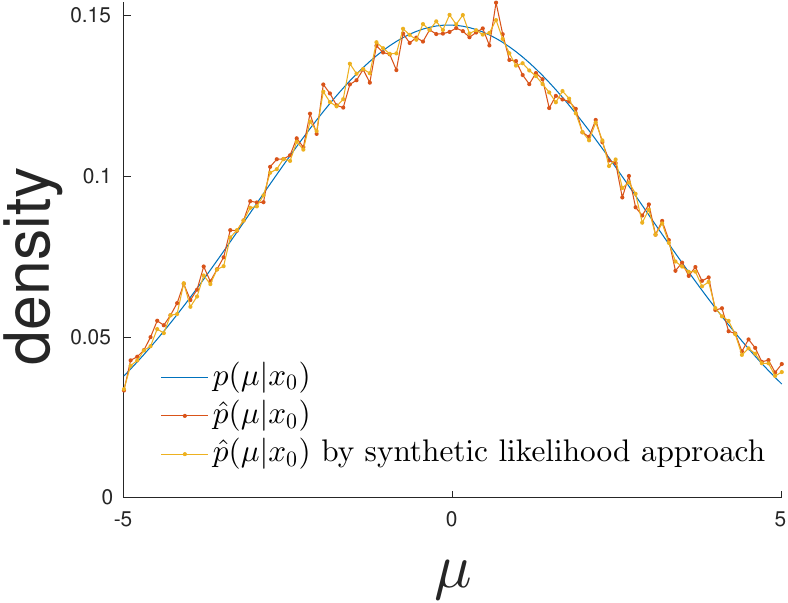}\label{fig:f1c}}
  \caption{Steps for estimating the posterior distribution of the mean
    of a Gaussian. (a) For any fixed value of $\mu$, $\lambdaMin$
    minimises the estimated prediction error $\mathcal{R}(\lambda)$
    (vertical line).  (b) The figure shows the true coefficients from
    Equation \eqref{eq:Gaussian} in black and the coefficients
    estimated by Algorithm~\ref{algo:approx_posterior} in colour. The
    algorithm sets the coefficients of unnecessary summary statistics
    automatically to zero. (c) Comparison of the estimated posterior
    with the posterior by the synthetic likelihood approach and the
    true posterior.}
\end{figure}

For Algorithm~\ref{algo:approx_posterior}, we used a ten-dimensional
summary statistic $\psi(x) = (1,x, \ldots,x^{b-1})$, with $b=10$, and
fixed $n_m = \ntheta = 1000$. As an illustration of step c of
Algorithm~\ref{algo:approx_posterior}, we show the prediction error
$\mathcal{R}(\lambda)$ in Figure~\ref{fig:f1a} as a function of
$\lambda$ for a fixed value of $\mu$. The chosen $\lambdaMin$
minimises the prediction error. Repeating step 3 in the algorithm for
different values of $\mu$ on a grid over the interval $[-5,5]$, we
estimated the ten-dimensional coefficient vector $\hat{\beta}(\mu)$ as
a function of $\mu$, which corresponds to an estimate
$\hat{\alpha}(\mu)$ of $\alpha(\mu)$, and hence of the posterior, by
Equation~\ref{eq:parametricEstimates}.

In Figure~\ref{fig:f1b}, we plot $\hat{\alpha}(\mu)$ and
$\alpha_0(\mu)$, $\alpha_1(\mu)$, $\alpha_2(\mu)$ from Equation
\eqref{eq:Gaussian} for $\mu \in [-5,5]$. We notice that the
estimated coefficients $\alpha_k$ are exactly zero for $k>2$ while
for $k\leq2$, they match the true coefficients up to random
fluctuations. This shows that our inference procedure can select the
summary statistics that are relevant for the estimation of the
posterior distribution from a larger set of candidates.

In Figure \ref{fig:f1c}, we compare the estimated posterior pdf
(yellow) with the true posterior pdf (blue). We can see that the
estimate matches the true posterior up to random fluctuations. The
figure further depicts the posterior obtained by the synthetic
likelihood approach of \citet{Wood_2010} (red) where the summary
statistics $\phi(x)$ are equal to $x$. Here, working with Gaussian
data, the performance of linear LFIRE by
penalised logistic regression and the performance of the existing
synthetic likelihood approach are practically equivalent.

\subsection{Autoregressive model with conditional heteroskedasticity}
\label{sec:ARCH}
In this example, the observed data are a time-series $\dataObs = \left(y^{(t)},
\ t=1,\ldots,T\right)$ produced by a lag-one autoregressive model
with conditional heteroskedasticity (ARCH(1)),
\begin{align}
  y^{(t)} &= \theta_1 y^{(t-1)} + e^{(t)}, & e^{(t)} & =  \xi^{(t)} \sqrt{ 0.2 + \theta_2
    (e^{(t-1)})^2},& t&=1,\ldots,T,& y^{(0)}&=0,
\label{eq:ARCH}
\end{align}
where $T = 100$, and $\xi^{(t)}$ and $e^{(0)}$ are independent
standard normal random variables. The parameters in the model,
$\theta_1$ and $\theta_2$, are correspondingly the mean and variance
process coefficients. The observed data were generated with $\theta^0
= (\theta^o_1,\theta^o_2)=(0.3,0.7)$ and we assume uniform priors
$\mathcal{U}(-1,1)$ and $\mathcal{U}(0,1)$ on the unknown parameters
$\theta_1$ and $\theta_2$, respectively. The true posterior
distribution of $\theta = (\theta_1,\theta_2)$ can be computed
numerically \cite[e.g.][Appendix 1.2.4]{Gutmann2018}. This enables us
to compare the estimated posterior with the true posterior using the
symmetrised Kullback-Leibler divergence ($\sKL$), where $\sKL$ between
two continuous distributions with densities $p$ and $q$ is defined as
\begin{eqnarray}
\label{eq:JS-div}
\sKL(p||q) = \frac{1}{2}\int{p(\theta)\log{\frac{p(\theta)}{q(\theta)}}d\theta} + \frac{1}{2}\int{q(\theta)\log{\frac{q(\theta)}{p(\theta)}}d\theta}.
\end{eqnarray}
Instead of comparing to the true posterior, one could compare to an
approximate posterior computed by conditioning on the observed value
of the summary statistics rather than the full data. We here focus on
the comparison to the true posterior in order to assess the overall
accuracy. The effect of the employed summary statistics is analysed in
\smref \ref{sec:arch_abc_comp}, and for intractable models considered
later in the paper, we construct reference posteriors via expensive
rejection ABC runs.

For estimating the posterior distribution with
Algorithm~\ref{algo:approx_posterior}, we used summary statistics
$\psi$ that measure the (nonlinear) temporal correlation between the
time-points, namely the auto-correlations with lag one up to five, all
pairwise combinations of them, and a constant. For checking the
robustness of the approach, we also considered the case where almost
$50\%$ of the summary statistics are irrelevant by augmenting the
above set of summary statistics by 15 white-noise random
variables. For synthetic likelihood, we used the auto-correlations as
the summary statistics without any additional irrelevant variables, as
the synthetic likelihood approach is typically not adapted to selecting
among relevant and irrelevant summary statistics. As explained in
Section~\ref{sec:selection}, synthetic likelihood always uses the
pairwise combinations of the summary statistics due to its underlying
Gaussianity assumption.
  
\begin{figure}
  \centering \subfloat[synthetic likelihood]{\includegraphics[width=0.33\textwidth]{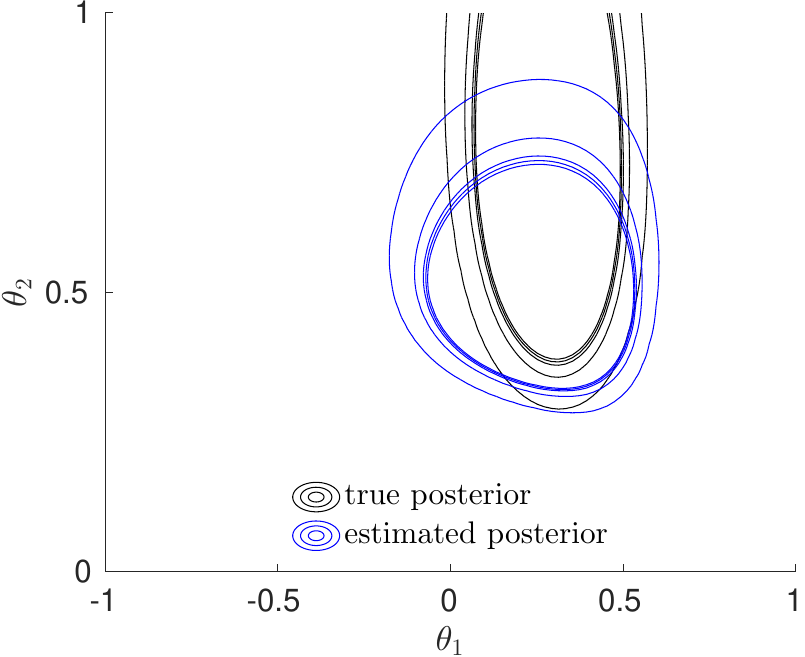}\label{fig:f2a}}
  \subfloat[linear LFIRE]{\includegraphics[width=0.33\textwidth]{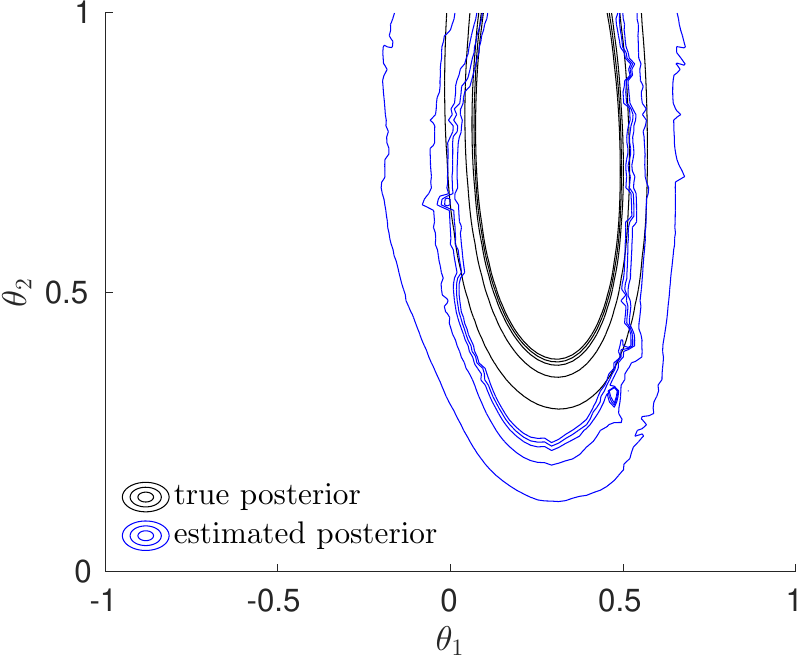}\label{fig:f2b}}
  \hfill \subfloat[linear LFIRE with irrelevant summaries]{\includegraphics[width=0.33\textwidth]{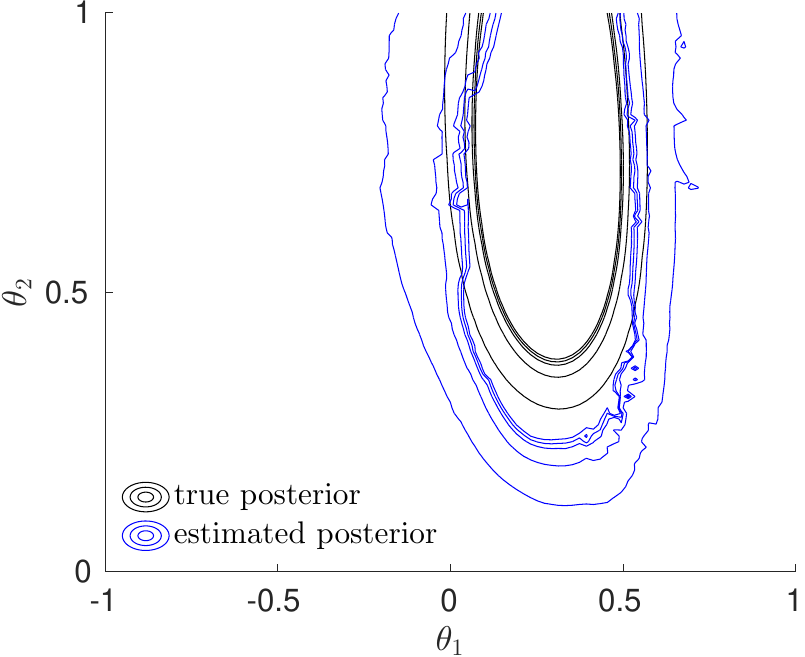}\label{fig:f2c}}
  \caption{ARCH(1): Contour plots of the posterior
    $\postHat(\theta|\dataObs)$ estimated by (a) synthetic likelihood,
    (b) the linear LFIRE method in Algorithm \ref{algo:approx_posterior}, and (c) linear LFIRE subject to 50\%
    irrelevant summary statistics. The range of the axes indicates the
    domain of the uniform prior. We used $n_{\theta} = n_{m} = 1000$
    for all results. The proposed linear LFIRE approach yields a
    better approximation than the synthetic likelihood approach and
    remains stable in the presence of irrelevant summary statistics.}
  \label{fig:contour_post_arch}
\end{figure}

We estimated the posterior distribution on a 100 by 100 mesh-grid over
the parameter space $[-1, 1]\times[0, 1]$ both for the proposed linear LFIRE and
the synthetic likelihood method. A comparison between two estimates is
shown in Figure~\ref{fig:contour_post_arch}. The figure shows that the
proposed approach yields a better approximation than the synthetic
likelihood approach. Moreover, the posterior estimated with our method
remains stable in the presence of the irrelevant summary
statistics. Our approximate posterior provides a reasonable
approximation to the exact posterior but we note that it has a larger
dispersion. The results in \smref \ref{sec:arch_abc_comp} suggest that
this difference is due to the summary statistics and not the inference method.

In order to assess the performance more systematically, we next
performed posterior inference for 100 observed time-series that were
each generated from Equation \eqref{eq:ARCH} with $\theta^0 =
(\theta^o_1,\theta^o_2)=(0.3,0.7)$. Table~\ref{tab:ARCH_comp} in \smref \ref{sec:appendix_tables} shows
the average value of the symmetrised Kullback-Leibler divergence for $\ntheta = n_m
\in \{100, 500, 1000\}$. The average divergence decreases as the
number of simulated data sets increases for our method, in contrast to
the synthetic likelihood approach. We can attribute the better performance of
our method to its ability to better handle non-Gaussian summary
statistics and its ability to select the summary statistics that are relevant.

We further compared the performance of linear LFIRE and
synthetic likelihood case-by-case for the 100 different observed data
sets. For this pairwise performance comparison, we computed the difference $\DeltaKL$
between the symmetrised Kullback-Leibler divergences
$\sKL(\hat{p}(\theta|\dataObs)||p(\theta|\dataObs))$ when
$\hat{p}(\theta|\dataObs)$ is estimated by the proposed method and by
synthetic likelihood. A value of $\DeltaKL < 0$ indicates a better
performance of the proposed method while a value $\DeltaKL > 0$
indicates that synthetic likelihood is performing better. As
$\DeltaKL$ depends on $\dataObs$, it is a random variable and we can
compute its empirical distribution on the 100 different inference
problems corresponding to different observed data sets. 

Figure~\ref{fig:arch_case_b_case} in \smref \ref{sec:appendix_figures} shows the distribution of $\DeltaKL$
when the irrelevant variables are absent (blue) and present (red) for
the proposed method. The area under the curve on the negative-side of
the x-axis is 82\% (irrelevant summaries absent) and 83\% (irrelevant summaries
present), which indicates a superior performance of the proposed
method over synthetic likelihood and robustness to the perturbing
irrelevant summary statistics. The p-values associated with a Wilcoxon
signed-rank test ($<10^{-10}$) demonstrate very strong evidence in
favour of the LFIRE method.

Figure \ref{fig:arch_scatter} in \smref \ref{sec:appendix_figures} shows a scatter plot of the symmetrised
Kullback-Leibler divergence for the LFIRE method and for the synthetic
likelihood. We see that the substantial majority of simulations fall
above the diagonal, indicating better performance of linear LFIRE compared to
synthetic likelihood, in line with the above findings.

\subsection{Comparison with a frequentist likelihood-ratio based method}
\label{sec:carl_comparison}
Here we compare the LFIRE method with a method based on approximating
likelihood ratios with calibrated discriminative classifiers
\citep[``carl'',][]{Cranmer_2015}, which provides an approximate
maximum likelihood estimator for a parameter $\theta$ by maximising
approximations of the ratio $p(x|\theta)/p(x | \theta_r)$, the ratio
of the freely parametrised likelihood $p(x | \theta)$ and the
likelihood evaluated at a reference value $\theta_r$, $p(x |
\theta_r)$. This is done by using a classifier to generate an
approximation to the ratio, followed by further calibration by use of
kernel density estimation. This corrective calibration step allows one
to use a wider range of loss function to train the
classifier (see the original paper for details).

The carl method relies on the choice of a reference $\theta_r$ to
construct the likelihood ratio. It is possible that a choice of
$\theta_r$ far from the true maximum likelihood estimate (MLE) will
provide a very large value of the likelihood ratio with
correspondingly high variance, making optimisation very
challenging. Even in a frequentist framework, we expect the LFIRE
methodology to be more robust to the choice of reference distribution,
since samples from the marginal distribution $p(x)$ will be generally
drawn from all regions covered by the prior $p(\theta)$. Consequently,
with the exception of the unlikely situation of very narrow and
mis-specified priors, the estimation of the ratio $p(x|\theta)/p(x)$
should be more stable.

\begin{figure}
  \centering
  \includegraphics[scale=0.6]{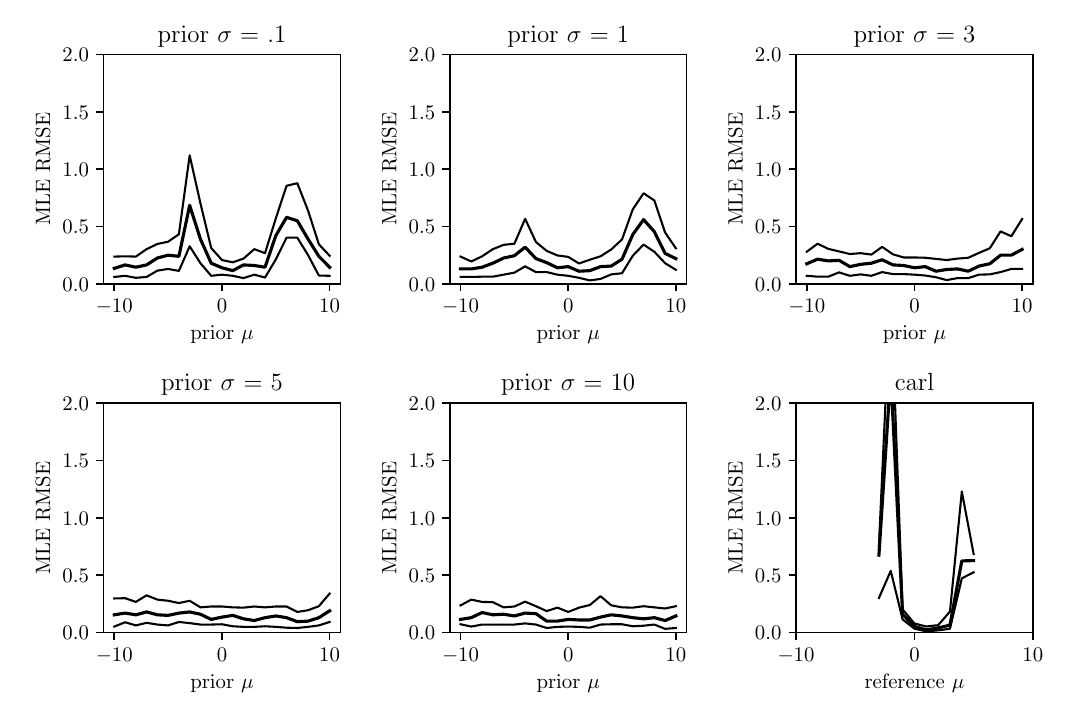}
    \caption{\label{fig:lfire_carl_comparison_MLE} Estimation of the
      mean of a univariate Gaussian, comparison of the root mean
      squared error (RMSE) of the approximate MLEs for LFIRE with
      different priors and carl with different reference points (bottom right).}
\end{figure}

\begin{figure}
  \hfill
  \includegraphics[scale=0.6]{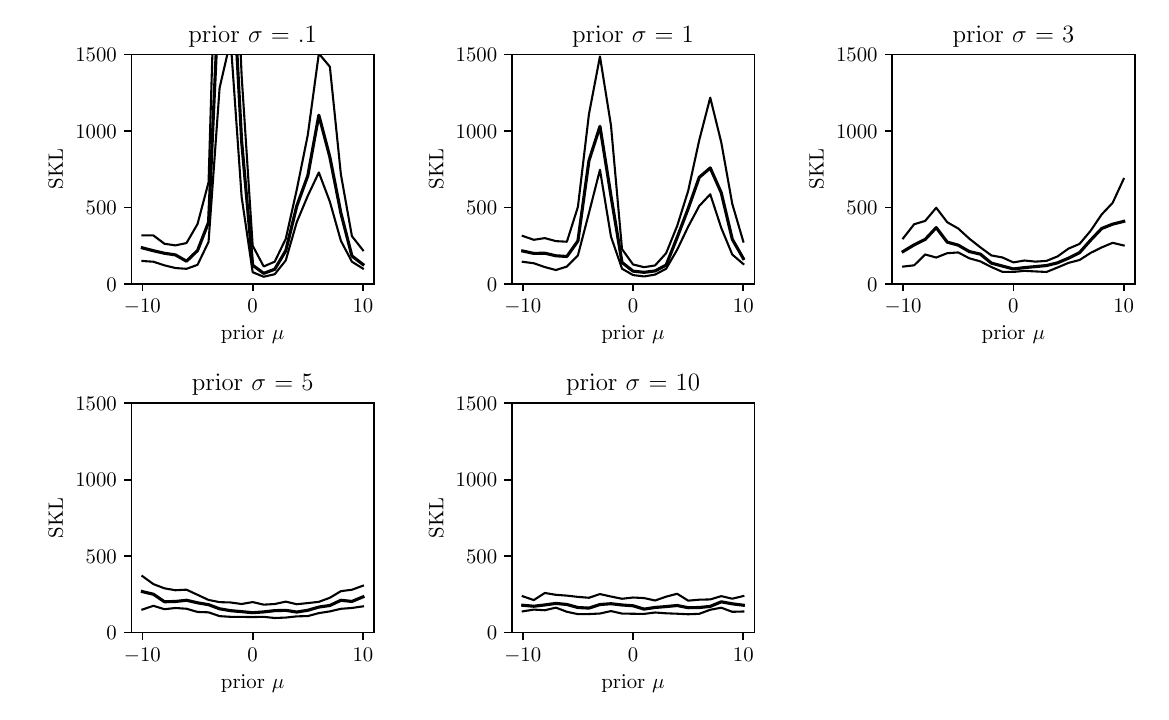}
    \caption{\label{fig:SKL_LFIRE} Setup as in Figure
      \ref{fig:lfire_carl_comparison_MLE} but assessing the posterior
      density estimates in terms of the symmetrised Kullback-Leibler
      ($\sKL$) distance to the true posteriors. The carl method did
      not lead to stable results and could not be included in the
      comparison.}
\end{figure}

We explore the behaviour of the two methods by estimating the mean of
a univariate Gaussian with known variance. Fifty data observations are
drawn from the true generative model with mean and variance equal to
one. The LFIRE method with $\ntheta=n_m=100$ was run with different
Gaussian prior distributions on the mean parameter, with prior
expectation varying between $-10$ and $10$, and prior standard
deviations taking values $[0.1, 1., 3., 5., 10.]$. The carl algorithm
of \citet{Cranmer_2015} was run using the associated software ``carl:
a likelihood free inference toolbox'', with reference parameter values
taking integer values from $-10$ to $10$. A multi-layer perceptron
global classifier was used for the carl algorithm, trained on $10 000$
simulation samples and parameter values drawn from the entire
parameter space. We found that increasing the number of simulations
did not improve performance of the carl algorithm. Both the
LFIRE and the carl simulations were repeated 50 times per
setup. Approximate MLEs were obtained for both methods: for carl, we
used the associated software package, for LFIRE,
they were computed by maximising the approximate likelihood in
\eqref{eq:nonparametricEstimates}.

Figure \ref{fig:lfire_carl_comparison_MLE} shows the root mean squared
errors (RMSE) of the obtained approximate MLEs, with the medians, 25th
and 75th quantiles over the 50 repetitions plotted. The carl method
(bottom right) led to small RMSEs when the reference point $\theta_r$
is well-chosen. When the reference point is further way from the true
parameter value, however, the RMSE becomes larger and when too far
away, the carl software failed and returned an uninformative default
value. LFIRE with a mis-specified overly confident prior (top left,
small standard deviation, prior mean far from the true value) produced
large RMSEs. For broader and more reasonable priors, LFIRE yielded
small RMSEs for a wide range of prior means, and was fairly robust to
the exact choice of the prior.

Figure \ref{fig:SKL_LFIRE} assesses the performance in posterior
density estimation in terms of the symmetrised Kullback-Leibler
divergence $\sKL$ between the approximate and true posterior. The
figure shows that LFIRE produced reasonable approximations unless the
prior was overly narrow and mis-specified. The carl method did not provide
computationally stable responses for the likelihood and hence
posterior for the entire parameter range considered, so it is not
included in the figure.

These empirical results are in line with the conceptual considerations
above. In particular, the use of the marginal $p(x)$ as a normalising
distribution in LFIRE leads to stable estimates across a broad range
of prior settings, which would be appropriate for a Bayesian
approach. By contrast, using a denominator likelihood conditioned on a
specific parameter value was found to be only accurate if the
reference value $\theta_r$ is close to the (unknown) true parameter,
but is unstable across an extended parameter range.

\section{Bayesian inference for nonlinear dynamical systems}
\label{sec:application}
We here apply linear LFIRE in Algorithm \ref{algo:approx_posterior} to two realistic
models with intractable likelihood functions and compare the inference
results with the results for the synthetic likelihood approach by
\citet{Wood_2010}. The first one is the ecological model of
\citet{Ricker_1954} that was also previously used by
\citet{Wood_2010}. The second one is the widely used weather
prediction model of \citet{Lorenz_1995} with a stochastic
reparametrisation \citep{Wilks_2005}, which we simply call ``Lorenz
model''. Both are time series models, and the inference is difficult
due to unobserved variables and their strongly nonlinear dynamics.

\subsection{Models}

\textbf{Ricker model.}  This is a model from ecology that describes the size of some animal population over time. The observed population size at time $t$, $y^{(t)}$, is assumed to be a stochastic observation
of the actual but unobservable population size $N^{(t)}$. Conditional on $N^{(t)}$, the observable $y^{(t)}$ is assumed Poisson distributed,
\begin{equation}
y^{(t)}|N^{(t)},\phi \sim \mathrm{Poisson}(\phi N^{(t)}),
\end{equation}
where $\phi$ is a scaling parameter. The dynamics of the unobservable population size 
$N^{(t)}$ is described by a stochastic version of the Ricker map \citep{Ricker_1954},
\begin{align}
\log{N^{(t)}} &= \log{r} + \log{N^{(t-1)}} - N^{(t-1)} + \sigma e^{(t)},& t &= 1,\ldots,T,& N^{(0)} &= 0,
\end{align}
where $T = 50$, $e^{(t)}$ are independent standard normal random
variables, $\log{r}$ is related to the log population growth rate, and
$\sigma$ is the standard deviation of the innovations. The model has
in total three parameters $\theta = (\log r, \sigma, \phi)$.  The
observed data $\dataObs$ are the time-series $(y^{(t)},
t=1,\ldots,T)$, generated using $\theta^0 = (\log r^0, \sigma^0,
\phi^0)=(3.8,0.3,10)$. We have assumed a uniform prior for all
parameters: $\mathcal{U}(3,5)$ for $\log{r}$, $\mathcal{U}(0,0.6)$ for $\sigma$, and
$\mathcal{U}(5,15)$ for $\phi$.

For our method, we use the set of 13 summary statistics $\phi$
suggested by \citet{Wood_2010} as well as all their pairwise
combinations and a constant in order to make the comparison with
synthetic likelihood fair -- as pointed out in Section
\ref{sec:selection}, synthetic likelihood implicitly uses the pairwise
combinations of the summary statistics due to its underlying
Gaussianity assumption. The set of 13 summary statistics $\phi$ are:
the mean observation $\bar{y}$, the number of zero observations,
auto-covariances with lag one up to five, the coefficients of a cubic
regression of the ordered differences $y^{(t)}-y^{(t-1)}$ on those of
the observed data, and the least squares estimates of the coefficients
for the model $(y^{(t+1)})^{0.3} =
b_1(y^{(t)})^{0.3}+b_2(y^{(t)})^{0.6}+\epsilon^{(t)}$, see
\citep{Wood_2010} for details.

\textbf{Lorenz model.} This model is a modification of the original
weather prediction model of \citet{Lorenz_1995} when fast weather
variables are unobserved \citep{Wilks_2005}. The model assumes
that weather stations measure a high-dimensional time-series of slow
weather variables $(y_k^{(t)}, k=1,\ldots,40)$, which follow a coupled
stochastic differential equation (SDE), called the forecast model
\citep{Wilks_2005},
\begin{align}
\label{eq:Lorenz}
 \frac{dy^{(t)}_k}{dt}&=-y^{(t)}_{k-1}(y^{(t)}_{k-2}-y^{(t)}_{k+1})-y_k^{(t)}+F-g(y_k^{(t)},\theta)+\eta_{k}^{(t)} \\
 g(y_k^{(t)},\theta) &= \theta_1 + \theta_2 y_k^{(t)},
\end{align}
where $\eta_{k}^{(t)}$ is stochastic and represents the uncertainty
due to the forcing of the unobserved fast weather variables. The
function $g(y_k^{(t)},\theta)$ represents the deterministic net effect
of the unobserved fast variables on the observable $y_k^{(t)},
k=1,\ldots,40$, and $F=10$. The model is cyclic in the variables
$y^{(t)}_k$, e.g.\ in Equation \eqref{eq:Lorenz} for $k=1$ we have $k-1
= 40$ and $k-2 = 39$. We assume that the initial values $y_k^{(0)},
k=1,\ldots, 40$ are known, and that the model is such that the time
interval $[0,4]$ corresponds to 20 days.

The above set of coupled SDEs does not have an analytical solution.
We discretised the 20 days time-interval $[0,4]$ into $T = 160$ equal
steps of $\Delta t = 0.025$, equivalent to 3 hours, and solved the
SDEs by using a 4th order Runge-Kutta solver at these time-points
\citep[Section 6.5]{Carnahan_1969}. In the discretised SDEs, following
\cite{Wilks_2005}, the stochastic forcing term is updated for an
interval of $\Delta t$ as
\begin{eqnarray*}
 \eta_{k}^{(t+\Delta t)} &= \phi \eta_{k}^{(t)} + \sqrt{1-\phi^2}e^{(t)}, \ t \in \{0, \Delta t, 2\Delta t, \ldots, 160\Delta t\},
\label{eq:stoch_forceterm_Lorenz}
\end{eqnarray*}
where the $e^{(t)}$ are independent standard normal random variables and $\eta^{(0)} = \sqrt{1-\phi^2}e^{(0)}$. 

The inference problem that we solve here is the estimation of the
posterior distribution of the parameters $\theta =
(\theta_1,\theta_2)$, called closure parameters in weather modelling,
from the 40 slow weather variables $y_k^{(t)}$, recorded over twenty
days. We simulated such observed data $\dataObs$ from the model by
solving the SDEs numerically as described above with $\theta^0 =
(\theta^o_1,\theta^o_2)=(2.0,0.1)$ over a period of twenty days. The uniform priors assumed for the parameters were $\mathcal{U}(0.5,3.5)$ for $\theta_1$ and $\mathcal{U}(0,0.3)$ for $\theta_2$. 

For the inference of the closure parameters $\theta$ of the Lorenz
model, \citet{Hakkarainen_2012} suggested six summary statistics: (1)
the mean of $y^{(t)}_k$, (2) the variance of $y^{(t)}_k$, (3) the
auto-co-variance of $y^{(t)}_k$ with time lag one, (4) the co-variance
of $y^{(t)}_k$ with its neighbour $y^{(t)}_{k+1}$, and (5, 6) the
cross-co-variance of $y^{(t)}_k$ with its two neighbours
$y^{(t)}_{k-1}$ and $y^{(t)}_{k+1}$ for time lag one. These values
were computed and averaged over all $k$ due to the symmetry in the
model. We used the six summary statistics for synthetic likelihood,
and, to make the comparison fair, for the proposed method, we also
used their pairwise combinations as well as a constant as in the
previous sections.

\subsection{Results}

\begin{figure}
  \centering
  \subfloat[]{\includegraphics[width=0.475\textwidth]{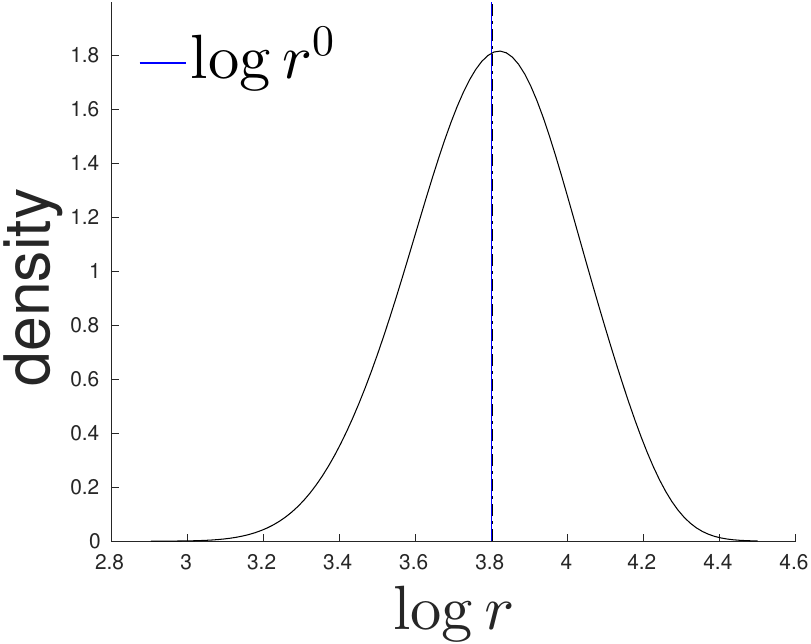}
  \label{fig:Ricker_logr_sigma_posterior}}\\
  \subfloat[]{\includegraphics[width=0.475\textwidth]{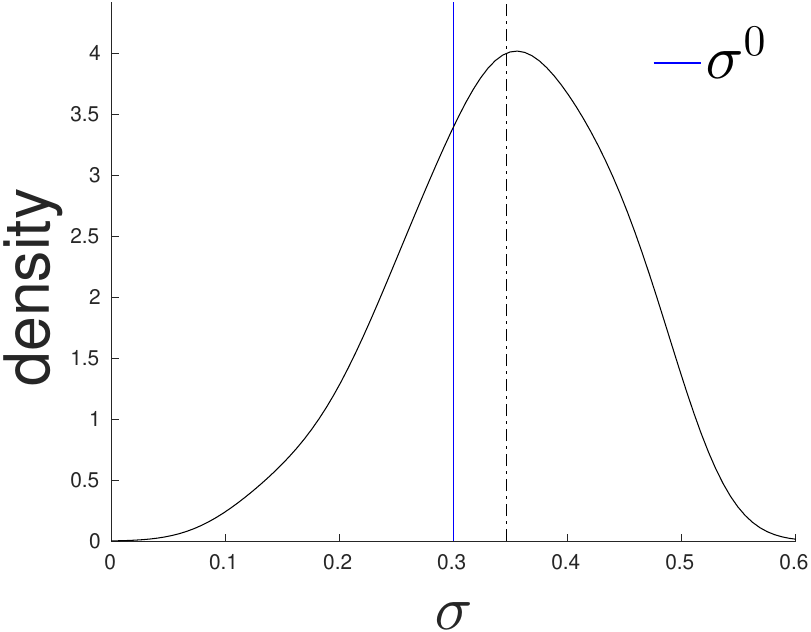}\label{fig:Ricker_sigma_phi_posterior}}
  \hfill
  \subfloat[]{\includegraphics[width=0.475\textwidth]{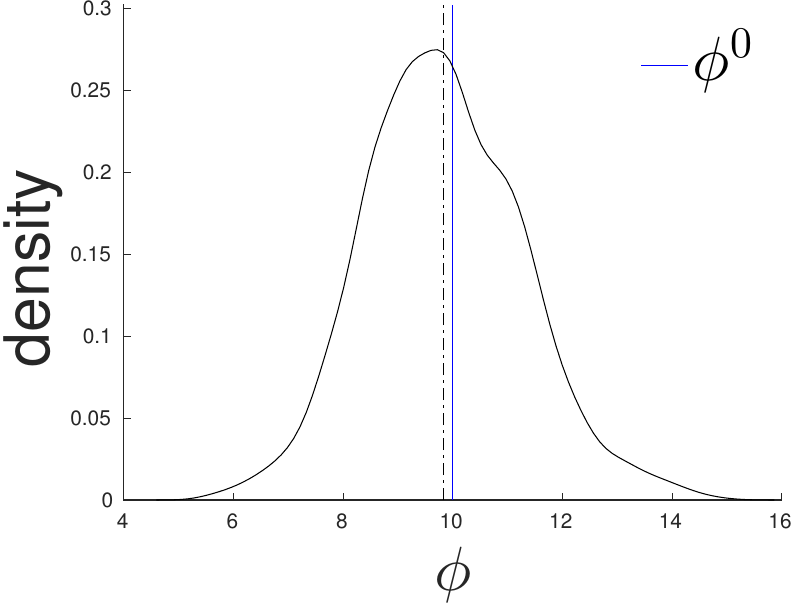}\label{fig:Ricker_logr_phi_posterior}}
\caption{Ricker model: Example marginal posterior distribution of (a)
  $\log r$, (b) $\sigma$ and (c) $\phi$, estimated with linear LFIRE
  in Algorithm \ref{algo:approx_posterior} using $\ntheta=n_m =
  100$. The blue vertical lines show the true parameter values $(\log
  r^0, \sigma^0, \phi^0)$ that we used to simulate the observed data
  and the black-dashed vertical lines show the corresponding estimated
  posterior means. The densities in (a--c) were estimated from
  posterior samples using a Gaussian kernel density estimator with
  bandwidths 0.1, 0.04, and 0.3, respectively.}
  \label{fig:Ricker_posterior}
\end{figure}

\begin{figure}
  \centering
  \includegraphics[width=0.6 \textwidth]{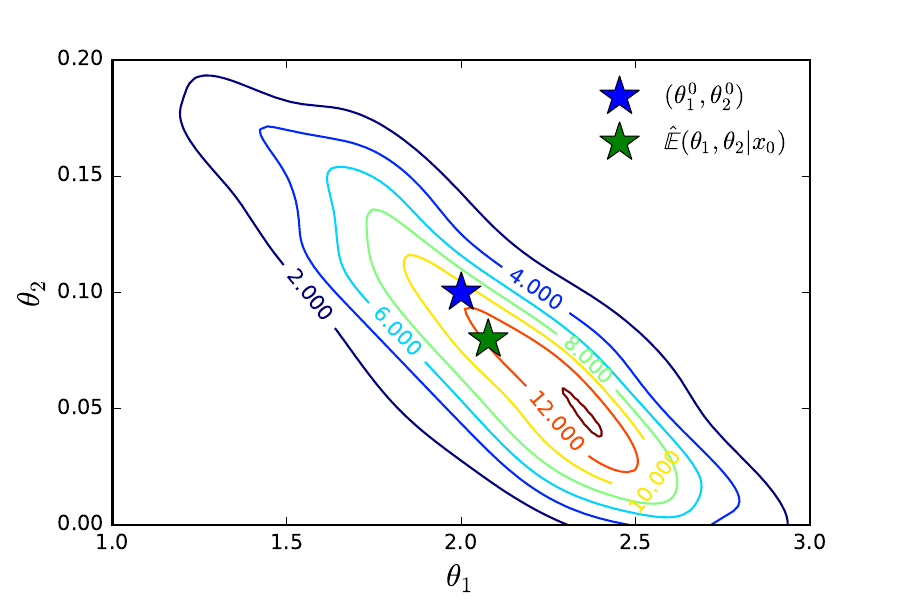}
  \caption{Lorenz model: Example posterior distribution of
    the closure parameters $(\theta_1,\theta_2)$ estimated with linear LFIRE in
    Algorithm \ref{algo:approx_posterior} using $\ntheta=n_m =
    100$. The blue and green asterisk indicate the true parameter values $(\theta_1^0,
    \theta_2^0)$ that were used to simulate the observed data and the estimated posterior mean of the parameters, respectively. The contour plot was generated from posterior samples by a weighted Gaussian kernel density estimator with bandwidth 0.5.}
  \label{fig:Lorenz_posterior}
\end{figure}

We used an importance sampling scheme \citep[IS]{Ripley_1987} by
sampling 10,000 samples from the prior distribution and computed their
weights using Algorithm \ref{algo:approx_posterior}, which is
equivalent to one generation of the SMC algorithm
\citep[SMC]{Cappe_2004, DelMoral_2006}. As suggested by \citet[see the
  method section in his paper]{Wood_2010}, for the synthetic
likelihood approach we used a robust variance-covariance matrix
estimation scheme for a better estimation of the likelihood
function. A simple approach is to add some scaled diagonal ``jitter'' to
the covariance matrix to ensure numerical stability when computing the
inverse.

Figure~\ref{fig:Ricker_posterior} shows example results for the Ricker
model, and Figure \ref{fig:Lorenz_posterior} example results for the
Lorenz model. While the results look reasonable, assessing their
accuracy rigorously is difficult due to the intractability of the
likelihood functions and the lack of ground truth posterior
distributions. We thus used the results from expensive rejection ABC
runs for reference (threshold set to achieve approximately 2\%
acceptance). We assessed the results in terms of the accuracy of the
posterior mean and posterior standard deviations.

The posterior mean $E_x[\hat{\theta}(x)]$ for linear LFIRE and
$E_x[\hat{\theta}_{SL}(x)]$ for the synthetic likelihood approach were
computed from the posterior samples. The relative errors of the
proposed approach and the synthetic likelihood were computed relative
to the ABC results for each element of the parameter vector $\theta$,
\begin{align}
 \relerr(x) &= \sqrt{\frac{ (E_x[\hat{\theta}(x)]-E_x[\hat{\theta}_{ABC}(x)])^2}{E_x[\hat{\theta}_{ABC}(x)]^2}}, &  \relerrsl(x) &= \sqrt{\frac{ (E_x[\hat{\theta}_{SL}(x)]-E_x[\hat{\theta}_{ABC}(x)])^2}{E_x[\hat{\theta}_{ABC}(x)]^2}}.
\end{align}
The squaring and division should be understood as element-wise
operations. As the relative error depends on the observed data, we
computed the error for $250$ different observed datasets
$\dataObs$. We performed a point-wise comparison between the proposed
method and synthetic likelihood by computing the difference
$\Deltarelerr$ between the relative errors for all elements in the
parameter vector $\theta$,
\begin{equation}
  \Deltarelerr = \relerr(\dataObs) - \relerrsl(\dataObs).
\label{eq:deltaMSE}
\end{equation}
Exactly the same procedure was used to assess the accuracy of the standard deviations.

For both the posterior means and standard deviations, a value of
$\Deltarelerr < 0$ means that the relative error for the proposed
method is smaller than the relative error for the synthetic likelihood
approach. A value of $\Deltarelerr > 0$, on the other hand, indicates
that the synthetic likelihood is performing better. As $\Deltarelerr$
is a function of $\dataObs$, we report the empirical distribution of
$\Deltarelerr$ computed from the $250$ different observed data sets
$\dataObs$.

Figures \ref{fig:Ricker_rel_error} to 
\ref{fig:Lorenz_rel_error_std} in \smref \ref{sec:appendix_figures} show
the empirical distribution of $\Deltarelerr$ for the posterior means and standard deviations for the Ricker and the Lorenz model.
All distributions are tilted toward negative values of $\Deltarelerr$
for all the parameters, which indicates that the proposed method is
generally performing better in both applications. As the proposed and
the synthetic likelihood method use exactly the same summary
statistics, we did not expect large improvements in the
performance. Nevertheless, the figures show that linear LFIRE
achieves better accuracy in the posterior mean for all but one
parameter where the performance is roughly equal, and better accuracy
in the posterior standard deviations in all cases. These results
correlate well with the findings for the ARCH model (note e.g.\ the
more accurate characterisation of the posterior uncertainty in Figure
\ref{fig:contour_post_arch}) and generally highlight the benefits
of LFIRE taking non-Gaussian properties of the summary statistics into
account.

We next analysed the impact of the improved inference on weather
prediction, which is the main area of application of the Lorenz
model. Having observed weather variables for $t \in [0,4]$, or 20
days, we would like to predict the weather of the next
days. We here consider prediction over a horizon of ten days, which
corresponds to $t \in [4,6]$.

Given $\dataObs$, we first estimated the posterior mean of the
parameters using the proposed and the synthetic likelihood
approach. Taking the final values of the observed data $(y_k^{(4)},
k=1,\ldots,40)$ as initial values, we then simulated the future
weather development using the SDE in Equation~\eqref{eq:Lorenz} for
both the true parameter value $\theta^0$, as well as for the two
competing sets of estimates. Let us denote the 40-dimensional time
series corresponding to $\theta^0$, $E_x[\hat{\theta}(x)]$ and
$E_x[\hat{\theta}_{SL}(x)]$ at time $t$ by $y^{(t)}$, $\hat{y}^{(t)}$,
and $\hat{y}_{\mathrm{SL}}^{(t)}$, respectively. We then compared the proposed
and the synthetic likelihood method by comparing their prediction
error. Denoting the Euclidean norm of a vector by $||.||$, we computed
\begin{eqnarray}
\zeta^{(t)}(\dataObs) = 
\frac{||y^{(t)}-\hat{y}_{\mathrm{SL}}^{(t)}||-||y^{(t)}-\hat{y}^{(t)}||}{||y^{(t)}-\hat{y}_{\mathrm{SL}}^{(t)}||}, \quad t \in (4,6],
\end{eqnarray}
which measures the relative decrease in the prediction error achieved
by the proposed method over synthetic likelihood. As the estimates
depend on the observed data $\dataObs$, $\zeta^{(t)}(\dataObs)$
depends on $\dataObs$. We assessed its distribution by computing its
values for 250 different $\dataObs$.

Figure~\ref{fig:lorenz_predic_err} in \smref \ref{sec:appendix_figures} shows the median, the
$1/4$ and the $3/4$ quantile of $\zeta^{(t)}(\dataObs)$ for $t
\in [4,6]$ corresponding to one to ten days in the future. We achieve
on average a clear improvement in prediction performance for the
first days; for longer-term forecasts, the improvement becomes
smaller, which is due to the inherent difficulty to make long-term
predictions for chaotic time series.

\section{Inference with high-dimensional summary statistics}
\label{sec:highdimsummstats}
Here we present the results of the LFIRE method applied to the
stochastic cell spreading model described in \citep{Price2017}. This
model is notable for its use of a large number of summary statistics
to determine the model parameters describing motility and
proliferation, $P_m$ and $P_p$. The summary statistics are the total
number of cells at the end of the experiment and the Hamming distances
between the image grids of cell populations evaluated at each time
point in the simulation, providing 145 summary statistics. This vector
was then combined with its own element-wise square and a constant,
resulting in a final total of 291 summary statistics.

\begin{figure}[tb!]
  \centering
  \subfloat[$\ntheta=n_m = 50$]{\includegraphics[width=0.475\textwidth]{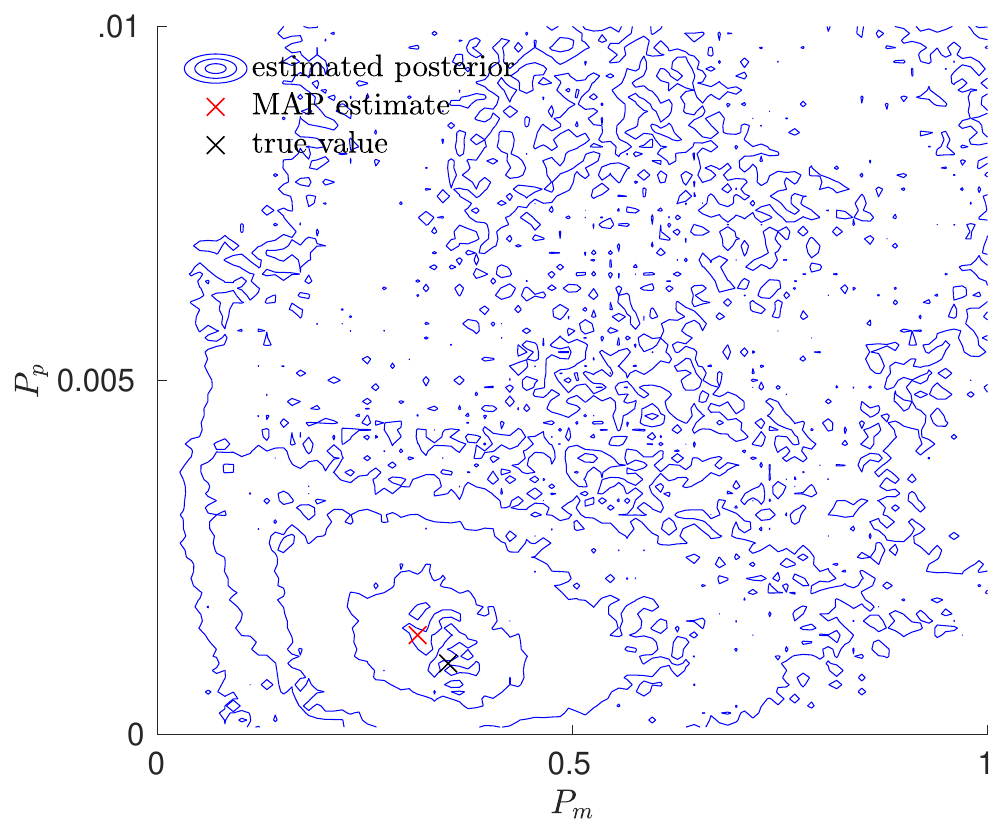}
  \label{fig:small_scratch_50}}\\
  \subfloat[$\ntheta=n_m = 100$]{\includegraphics[width=0.475\textwidth]{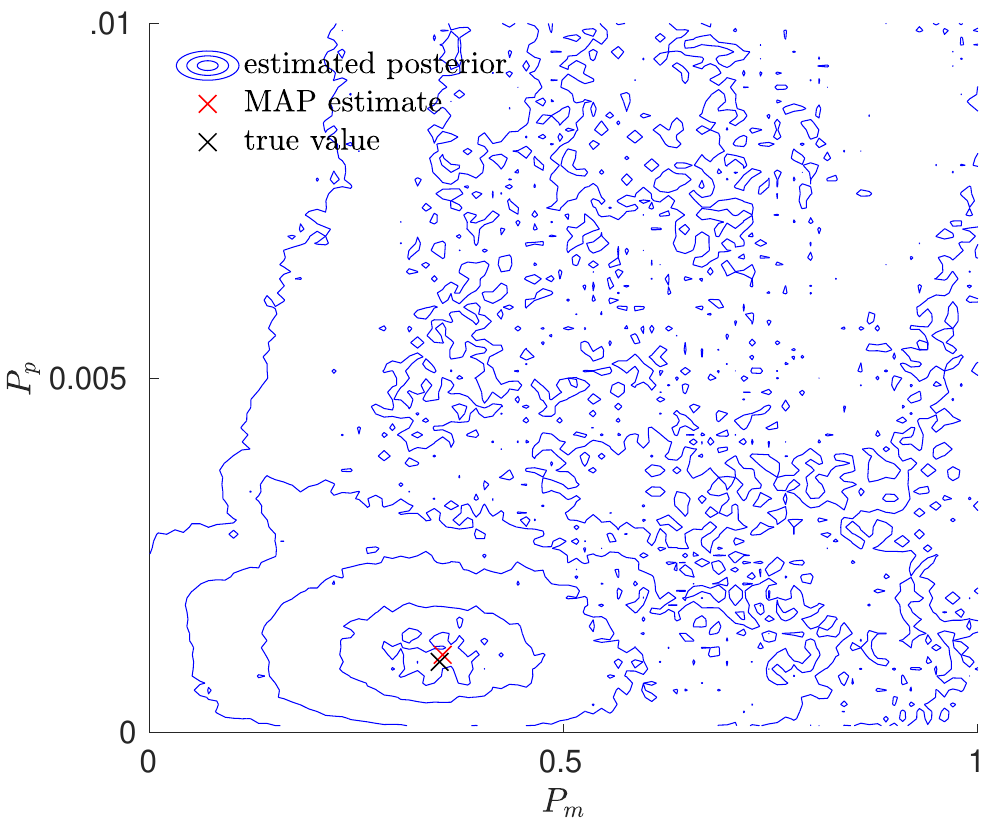}
  \label{fig:small_scratch_100}}
  \hfill
  \subfloat[$\ntheta=n_m = 150$]{\includegraphics[width=0.475\textwidth]{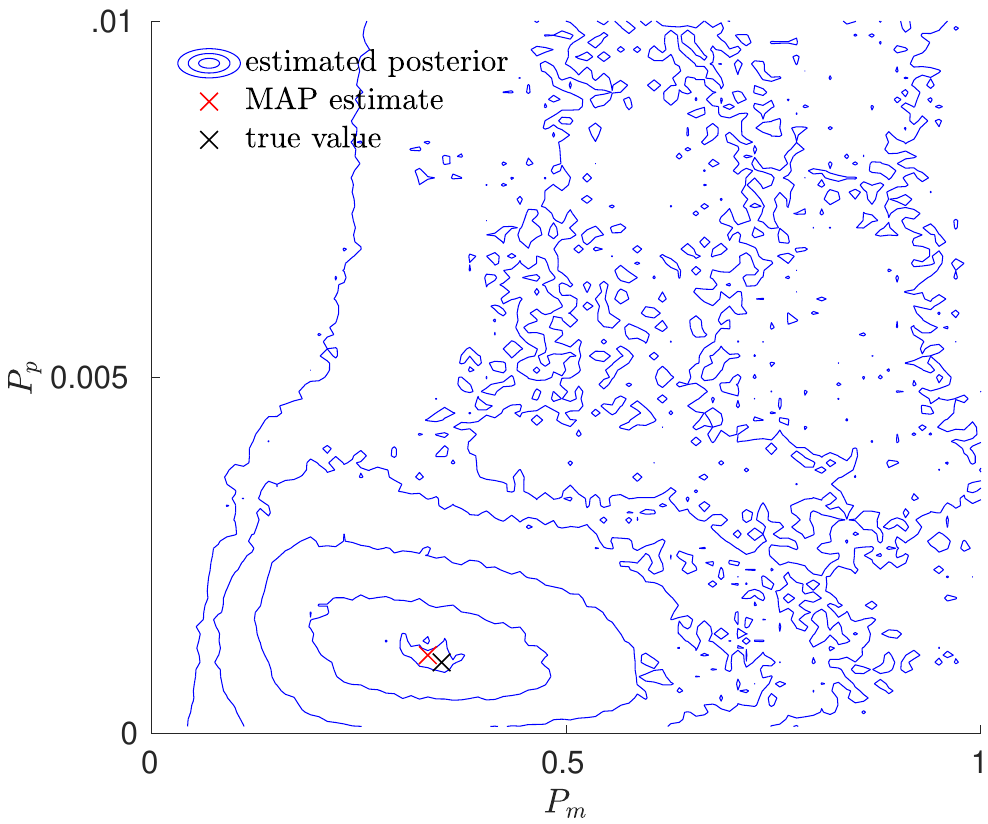}
  \label{fig:small_scratch_150}}
\caption{Cell spreading model: Contour plots of the approximate
  posterior computed by linear LFIRE in Algorithm
  \ref{algo:approx_posterior} for the parameters $P_m$ and $P_p$. Each
  panel corresponds to different numbers of simulated data points
  $\ntheta=n_m$ to train the classifier. The true values and MAP
  estimates of the parameters are also displayed in the plots.}
  \label{fig:small_scratch_posteriors}
\end{figure}

The linear LFIRE method using a lasso-type regularisation is well-positioned
to perform efficient inference for such a model, as it can select the
summary statistics that are most informative for the characterisation
of the posterior distribution. We performed inference with true values
of $P_m=0.35$ and $P_p=0.001$, and varied the amount of simulated data
used to train the classifier, with values of $\ntheta=n_m \in \{50,
100, 150\}$. Given the prior knowledge that the $P_p$ would take small
values, we asserted uniform priors over each model parameter between
$[0,1]$ and $[0, 0.01]$, respectively.

The results are presented in Figure
\ref{fig:small_scratch_posteriors}. It is seen that the posterior
becomes more stably characterised as $\ntheta=n_m$ increases from 50 to 150, with the
MAP estimates clearly improving with the extra training data.
The results are in line with those presented by \cite{Price2017} who
used synthetic likelihood, but employing a much larger number of
simulations (namely $\ntheta \in \{2500, 5000, 10000\}$). We were not
able to make synthetic likelihood work with the low numbers of
$\ntheta$ used in LFIRE but recent work on synthetic likelihood
\citep{Ong2018} show that shrinkage estimation methods enable values
of $\ntheta \in \{500, 1000\}$.

We further compared the number of non-zero summary statistic weights
that linear LFIRE uses for each value of $\ntheta=n_m$. As the classifier is
exposed to more training data, we would expect it to select more
summary statistics as more evidence becomes available. This phenomenon
is observed in our simulations, with the $\ntheta=n_m=50$, $100$ and
$150$ simulations selecting an average of $17.3$, $23.9$ and $30.5$
summary statistics, respectively, from a total number of size
$291$. This demonstrates that the method is able to both select from
a large pool of summary statistics and form a more complex
classification model when more computational resources are made
available.

\section{Computational aspects}
\label{sec:timing}
In this section we discuss the computational cost of LFIRE and compare
it with the existing synthetic likelihood approach.

Both methods require generating the data set $\Xtheta$, which most
often will dominate the computational cost. Linear LFIRE has the
additional cost of constructing the set $\Xm$ once, and the cost of
performing penalised logistic regression for each $\theta$, including
potentially computationally intensive cross-validation to establish
the regularisation strength. Synthetic likelihood, on the other hand,
requires inversion of the covariance matrix of the summary statistics
for each $\theta$.  Cross-validation can also be used for
regularisation of the synthetic likelihood with a graphical lasso for
robust inference in situations with a large or poorly-conditioned
covariance matrix \citep{An2019}. When the covariance matrix is
well-conditioned, penalised logistic regression with cross-validation
is more expensive than standard inversion of the covariance matrix,
and the difference in computational cost can be seen as the price that
needs to be paid for the relaxation of the Gaussianity assumption, and
for feature selection through regularised inference. If, however,
simulating data from the model is in the computational bottleneck, the
extra cost of regularised logistic regression causes comparably little
overhead.

We support these considerations with timing data that were collected
for the ARCH and the cell spreading (``scratch'') model. Since the absolute
computational times are platform-specific, we consider the relative
amount of time spent performing simulations and posterior
estimation. Simulation times were averaged for robustness over one
million simulations, with parameters drawn from the uniform priors
used in the experiments described in Sections \ref{sec:ARCH} and
\ref{sec:highdimsummstats}. Similarly, posterior estimation times were
averaged over 100 evaluations of posterior proxies, including
performing penalised logistic regression through cross-validation,
defined over a grid across the uniform prior. The relative balance of
computational times between simulations and posterior estimation
considered 100 evaluations of the posterior proxies, with $\ntheta$
simulated data sets used for each evaluation. Parallel computational
resources were not considered in the analysis: they could definitely
affect the relative computational times, but with a heavy dependence
on the local computing platform and specific inference procedure.

Since both of the ARCH and cell spreading model are computationally
inexpensive, we also consider a hypothetical model for which
simulations take one second, which is still rather cheap. Likelihood
proxy evaluation times were assumed to be 0.09 seconds for the
synthetic likelihood and 10 seconds for LFIRE as approximate midpoints
of those observed for the other models.

Table \ref{Tab:timingproportions} presents the proportion of
computational time spent performing simulations for different
simulator models, posterior approximation methods and values of
$\ntheta$. We see that for the very cheap ARCH simulations, the LFIRE method spends a
majority of its time performing posterior estimation. For the
moderately more expensive cell spreading (``scratch'') simulations, posterior estimation
computations are still dominant, but the relative costs become more balanced. For the hypothetical (but not
unrealistic) one second simulator, we see that a comfortable majority
of computational time is now spent performing simulations for both
linear LFIRE and synthetic likelihood (SL).

In summary, we see that while the LFIRE method requires more time than
synthetic likelihood for each posterior estimate, for simulators with
non-trivial computational demands the proportion of time spent on
generating data is dominant for both methods.

\begin{table}[t!]
\centering
 \begin{tabular}{c c c c c c c c c c c}\toprule
   \multicolumn{3}{c}{ARCH}& \phantom{a}& \multicolumn{3}{c}{Scratch}&\phantom{a}&\multicolumn{3}{c}{Hypothetical 1s simulator}\\
   \cmidrule{1-3}  \cmidrule{5-7}  \cmidrule{9-11}  
 $n_\theta$&LFIRE&SL&& $n_\theta$&LFIRE&SL&& $n_\theta$&LFIRE&SL\\\midrule
 $100$&0.0295&0.4234 &&$50$&0.1623&0.3177 &&$50$&0.8347&0.9982\\
  $500$&0.0171&0.7859 &&$100$&0.1738&0.4822 &&$100$&0.9099&0.9991\\
 &&&&$150$&0.1231&0.5828 &&$150$&0.9381&0.9994\\
 \bottomrule
\end{tabular}
\caption{Proportion of total compute time dedicated to simulation.}
\label{Tab:timingproportions}
\end{table}

\section{Discussion}
We considered the problem of estimating the posterior
density when the likelihood function is intractable but generating
data from the model is possible. We framed the posterior density
estimation problem as a density ratio estimation problem. The latter
problem can be solved by (nonlinear) logistic regression and is thus
related to classification and contrastive learning.

This approach for posterior estimation with generative models mirrors
the approach of \citet{Gutmann_2012} for the estimation of
unnormalised models. The main difference is that here, as well as in
the related work by \citet{Pham_2014, Cranmer_2015}, we classify
between two simulated data sets while \citet{Gutmann_2012} classified
between the observed data and simulated reference data. This
difference reflects the fact that generating samples is relatively
easy for generative models while typically difficult for unnormalised
models. As we are guaranteed to have enough data to train the
classifier, the main advantage of working with two simulated data sets
is that it supports posterior inference given a single observed datum
only.

Our approach requires that several samples from the model are generated
for the estimation of the posterior, like for synthetic likelihood
\citep{Wood_2010}. While the sampling can be performed perfectly in
parallel, it constitutes the main computational cost unless the model
is very cheap to simulate. There are several ways to reduce the
inference cost: First, Bayesian optimisation can be used to
intelligently decide where to evaluate the posterior as previously
done for the synthetic likelihood, thus reducing unnecessary
computations \citep{Gutmann_2016, Jarvenpaa2018b}. Second, rather than
pointwise estimation, the inference can be amortised with respect to the
parameters \citep{Hermans2020} or one can learn the relation between
the parameters and the log-ratio from already computed
parameter-ratio pairs. An initial estimate of the posterior can
thereby be obtained without any new sampling from the model, and
additional computations may only be spent on fine-tuning that
estimate. Third, for prior distributions much broader than the
posterior, performing logistic regression with samples from the
marginal distribution is not very efficient. Iteratively constructing
a proposal distribution that is closer to the posterior, will likely
lead to computational gains. Finally, most computations can be
performed offline before the observed data are seen, so that
computations can be cached and recycled for newly observed data sets,
which reduces the effective cost and enables amortised inference and
``crowd-sourcing'' of computations. This kind of (shared)
pre-computations can be particularly advantageous when the posterior
needs to be estimated as part of a decision making process that is
subject to time constraints.

A key feature of the proposed method is the automated selection and
combination of summary statistics from a large pool of candidates.
While there are several works on summary statistics selection in the
framework of approximate Bayesian computation \citep{Aeschbacher_2012,
  Fearnhead_2012, Blum_2013, Gutmann_2014, Marin_2016, Gutmann2018,
  Jiang2018}, there is comparably little corresponding work on
synthetic likelihood \citep{Wood_2010} with the exception of the
recent work by \citet{An2019} and \citet{Ong2018} whose robust estimation
techniques of the (inverse) covariance matrix is broadly related to
summary statistics selection. We have shown that synthetic likelihood
is a special case of the proposed approach so that our techniques for
summary statistics selection could also be used there. 

While the cited methods for summary statistics selection in ABC might
be adaptable for use with synthetic likelihood, the summary statistics
generally have to be transformed before use, in order to match the
multivariate Gaussianity assumption of synthetic likelihood. Finding
such a joint transformation of summaries to fulfil the multivariate
Gaussianity criterion is generally challenging, as it is very
difficult to constrain the resulting multivariate distribution's
higher-order moments, e.g. the co-skewness and co-kurtosis, without
losing information. However, it is always possible to average across
multiple evaluations of summary statistics and use a central limit
theorem to asymptotically approach a multivariate Gaussian
distribution. This is in contrast to our approach that automatically
adapts to non-Gaussianity of the summary statistics.

Our results showed that the proposed method can effectively select
relevant summary statistics. We used the method to remove completely
irrelevant ones but also to adaptively include more relevant ones
when more computational resources become available. Moreover, the
ability to automatically select data summaries substantially
contributes to the interpretability of the inference procedure, and
the selected summary statistics may provide additional insights into
the fundamental scientific question at hand.
While automated selection of summary statistics from a large pool of
candidates alleviates the burden on the user to provide carefully
engineered summary statistics it assumes that some of the candidates
are suitable in the first place. The intrinsic connection of the
proposed approach to classification facilitates the learning of
summaries from raw data \citep{Dinev2018} thereby partly addressing
this point.

We have seen that the proposed approach can well handle
high-dimensional summary statistics. The separate problem of
likelihood-free inference for high-dimensional parameter spaces is a
highly relevant question. This is a very challenging problem without a
generally accepted solution: the LFIRE methodology might perform well
in this context, because it does not involve the choice of an
acceptance threshold or other kernel as in typical ABC, which is often
a problem when inferring high-dimensional parameters. However, it was
not developed with such problems specifically in mind, and such an
investigation does not fall within the scope of this paper.

We have used a linear basis expansion and logistic regression to
implement the proposed framework of likelihood-free inference by ratio
estimation. While more general regression models and other loss
functions such as Bregman divergences \citep{Gutmann2011b,
  Sugiyama_2012} can be used, we found that already this simple
instance of the framework provided a generalisation of the synthetic
likelihood approach with typically more accurate estimation results.

Our findings suggest that likelihood-free inference by ratio
estimation is a useful technique, and the proposed rich \emph{framework} opens up
several directions to new inference methods based on logistic
regression or other density ratio estimation schemes that can be used
whenever the likelihood function is not available but sampling from
the model is possible.

\clearpage

\ifsepsuppl
\else
\clearpage
\renewcommand\appendixname{Supplementary Material}
\appendix
\section{Proof of Equation \eqref{eq:hoptimal}}
\label{proof:logr-opt}
We here prove that $\log r(x,\theta) = \log\left(\pmodel(x|\theta) /
\pmarg(x) \right)$ minimises $\J(h,\theta)$ in Equation
\eqref{eq:Jthetah} in the limit of large $\ntheta$ and $n_m$.

We first simplify the notation and denote $\xtheta$ by $x$, its pdf
$p(x | \theta)$ by $\pdata$, $\ntheta$ by $n$, $\xm$ by $y$, its pdf
$p(x)$ by $\pnoise$, and $n_m$ by $m$. Moreover, as $\theta$ is
considered fixed for this step, we drop the dependency of $\J$ on
$\theta$. Equation \eqref{eq:Jthetah} thus reads
\begin{equation}
\J(h) = \frac{1}{n+m} \left\{ \sum_{i=1}^{n}
\log\left[1+\nu \exp(-h(x_i))\right] + \sum_{i=1}^{m}
\log\left[1+\frac{1}{\nu} \exp(h(y_i))\right] \right\}.
\end{equation}
We will consider the limit where $n$ and $m$ are large, with fixed ratio $\nu = m/n$. For that purpose we write $\J$ as
\begin{align}
\J(h) &= \frac{n}{n+m} \left\{\frac{1}{n} \sum_{i=1}^{n}
\log\left[1+\nu \exp(-h(x_i))\right] + \frac{1}{n}\sum_{i=1}^{m}
\log\left[1+\frac{1}{\nu} \exp(h(y_i))\right] \right\}\\
&= \frac{n}{n+m} \left\{\frac{1}{n} \sum_{i=1}^{n}
\log\left[1+\nu \exp(-h(x_i))\right] + \nu \frac{1}{m}\sum_{i=1}^{m}
\log\left[1+\frac{1}{\nu} \exp(h(y_i))\right] \right\}\\
&= \frac{1}{1+\nu} \left\{\frac{1}{n} \sum_{i=1}^{n}
\log\left[1+\nu \exp(-h(x_i))\right] + \nu \frac{1}{m}\sum_{i=1}^{m}
\log\left[1+\frac{1}{\nu} \exp(h(y_i))\right] \right\}
\end{align}
In the stated limit, $\J(h)$ thus equals $\J(h) = \Jbar(h) / (1+\nu)$, where
\begin{align}
\Jbar(h) &=  \E_x \log\left[1+\nu \exp (-h(x))\right] + \nu \E_y \log\left[1+\frac{1}{\nu} \exp(h(y))\right].
\end{align}
The function $h^*$ that minimises $\Jbar(h)$ also minimises $\J(h)$ in the limit of large $n$ and $m$. To determine $h^*$ we apply
\begin{align}
\log\left(1+\frac{1}{\nu} \exp h\right) & = \log(\nu \exp(-h)+1) -\log(\nu \exp(-h))
\end{align}
and re-write $\Jbar$ as
\begin{align}
\Jbar(h) =&  \E_x \log(1+\nu \exp (-h(x))) + \nu \E_y \log(\nu \exp(-h(y))+1)  \nonumber \\
&- \nu \E_y \log(\nu \exp(-h(y)))  \\
=& \E_x \log(1+\nu \exp (-h(x))) + \nu \E_y \log \left(1+\nu \exp(-h(y))\right) \nonumber \\&- \nu \log \nu + \nu \E_y h(y)  \\
=& \int \pdata(u) \log(1+\nu \exp(-h(u))) du +\nu \int \pnoise(u) \log(1+\nu \exp(-h(u)))du \nonumber \\
& - \nu \log \nu + \nu \int \pnoise(u) h(u)du \\
=& \int \left(\pdata(u)+\nu \pnoise(u)\right) \log(1+\nu \exp(-h(u))) du  \nonumber \\&- \nu \log \nu+ \nu \int \pnoise(u) h(u) du.
\end{align}
We now expand $\Jbar(h+\epsilon q)$ around $h$ for an arbitrary function $q$ and a small scalar $\epsilon$.
With
\begin{align}
\log(1+\nu \exp\left(-h(u)-\epsilon q(u)\right)) =&  \log(1+\nu \exp(-h(u)) \nonumber \\
& -\epsilon q(u) \frac{\nu \exp(-h(u))}{1+\nu \exp(-h(u))} \nonumber \\
&+ \frac{\epsilon^2 q(u)^2}{2}\frac{\nu \exp(-h(u))}{1+\nu \exp(-h(u))}\frac{1}{1+\nu \exp(-h(u))} \nonumber \\
&+O(\epsilon^3)
\end{align}
we have
\begin{align}
\Jbar(h+\epsilon q) = & \int \left(\pdata(u)+\nu \pnoise(u)\right) \log(1+\nu \exp(-h(u)) du \nonumber\\
& - \int \left(\pdata(u)+\nu \pnoise(u)\right) \epsilon q(u) \frac{\nu \exp(-h(u))}{1+\nu \exp(-h(u))} du \nonumber \\
& +\int \left(\pdata(u)+\nu \pnoise(u)\right) \frac{\epsilon^2 q(u)^2}{2}\frac{\nu \exp(-h(u))}{1+\nu \exp(-h(u))}\frac{1}{1+\nu \exp(-h(u))} du \nonumber \\
& - \nu \log \nu + \nu \int \pnoise(u) h(u) du + \nu \int \pnoise(u) \epsilon q(u) du +O(\epsilon^3).
\end{align}
\vspace{-1ex}
Collecting terms gives
\begin{align}
\Jbar(h+\epsilon q) = & \Jbar(h) - \epsilon \int q(u) \left(\left(\pdata(u)+\nu \pnoise(u)\right) \frac{\nu \exp(-h(u))}{1+\nu
	\exp(-h(u))}-\nu \pnoise(u) \right) du \nonumber \\
& +\frac{\epsilon^2}{2}
\int q(u)^2 \left(\pdata(u)+\nu \pnoise(u)\right) \frac{\nu
	\exp(-h(u))}{1+\nu \exp(-h(u))}\frac{1}{1+\nu \exp(-h(u))} du \nonumber \\ 
&  +O(\epsilon^3).
\end{align}
The second-order term is positive for all (non-trivial) $q$ and $h$. The first-order term is zero for all $q$ if and only if
\begin{align}
\nu \pnoise(u) &= \frac{\pdata (u)+\nu \pnoise(u)}{1+\frac{1}{\nu}\exp(h^*(u))} & \Leftrightarrow && \nu \pnoise(u) + \pnoise(u)\exp(h^*(u)) &=\pdata (u)+\nu \pnoise(u) 
\end{align}
that is, if and only if
\begin{align}
\exp(h^*(u)) &= \frac{\pdata(u)}{\pnoise(u)},
\end{align}
which shows that $h^* = \log\left(\pdata/\pnoise\right)$ minimises
$\Jbar$. With the notation from the main text, $h^* = \log \left(
\pmodel(x|\theta) / \pmarg(x) \right)$, which equals $\log
r(x,\theta)$, and thus proves the claim. Note that the same ratio is
obtained for $\pdata(u) = p(x | \theta) f(\theta)$ and $\pnoise(u) =
p(x)f(\theta)$ because $f(\theta)$ cancels out. Here, $f(\theta)$ can
be any density with support on the parameter space where we want to
evaluate the ratio or posterior, and the classification would be
performed in the joint $\theta, x$ space.

\section{ARCH model: effect of summary statistics}
\label{sec:arch_abc_comp}
Unless the summary statistics are sufficient, the posteriors
conditioned on the observed data and the posteriors conditioned on the
observed summary statistics are different. In the main text, we
performed an overall comparison between the approximate and exact
posteriors. This is valuable because it measures what we ultimately
care about. But it confounds the effect of the summary statistics and
the effect of the ratio estimation approach. In order to separate the
two effects, we here present an additional comparison using a
``gold-standard'' rejection ABC algorithm with a small ($2.94\cdot
10^{-4}$) rejection threshold, drawing 1000 samples for each of the
100 simulated data sets, which provide samples from the posterior
conditioned on the summary statistics. We then directly compared the
posterior means and standard deviations of the ABC, linear LFIRE, and the
exact posteriors.

Averaged over observed data sets, the ABC algorithm yielded posterior
means of $\theta_{ABC}=[0.2723, 0.6345]$, and average posterior
standard deviations took values of $0.1728$ and $0.1783$ for each
parameter. Performing quadrature over the grid of parameter values for
the LFIRE simulations, then averaging over all 100 observed data sets
gave mean estimates of $[0.3038, 0.6159]$ and standard deviation
estimates of $[0.1494 , 0.1928]$, and a similar quadrature approach
for the true posterior conditioned on the whole data set gave values
of $[0.2924, 0.6779]$ with average standard deviations of $[0.0921,
0.1510]$. We observe that the posterior means of all these
distributions are similar, but the standard deviations of $\theta_1$
differ by a factor of approximately two between the ABC samples and
the true posterior, in line with the broader posterior reported in
the main text. The summary statistics thus broaden the posterior,
which also means, because the posteriors integrate to one, that the
estimated ratio $\hat{r}(x, \theta)$ is typically smaller than the
true ratio.

\newpage

\section{Supplementary Table}
\label{sec:appendix_tables}

\begin{table}[!ht]
	\begin{center}
		\begin{tabular}{lrrr}\toprule
			Method & $n_s$ = 100 & $n_s$ = 500 & $n_s$ = 1000 \\ \midrule
			Synthetic likelihood & 1.82  & 1.80 & 2.25 \\ 
			Linear LFIRE  & 2.04  & 1.57 & 1.48 \\ 
			Linear LFIRE with irrelevant summaries &  3.24  & 1.60 & 1.51\\
			\bottomrule
		\end{tabular}
	\end{center}
	\caption{\label{tab:ARCH_comp} ARCH(1): Average symmetrised Kullback-Leibler divergence between the true and estimated posterior for $\ntheta = n_m = n_s \in \{100, 500, 1000\}$. Smaller values of the divergence mean better results.}
\end{table}

\section{Supplementary Figures}
\label{sec:appendix_figures}
\vspace{4ex}
\begin{figure}[h!]
	\centering
	\includegraphics[width = 0.7\textwidth]{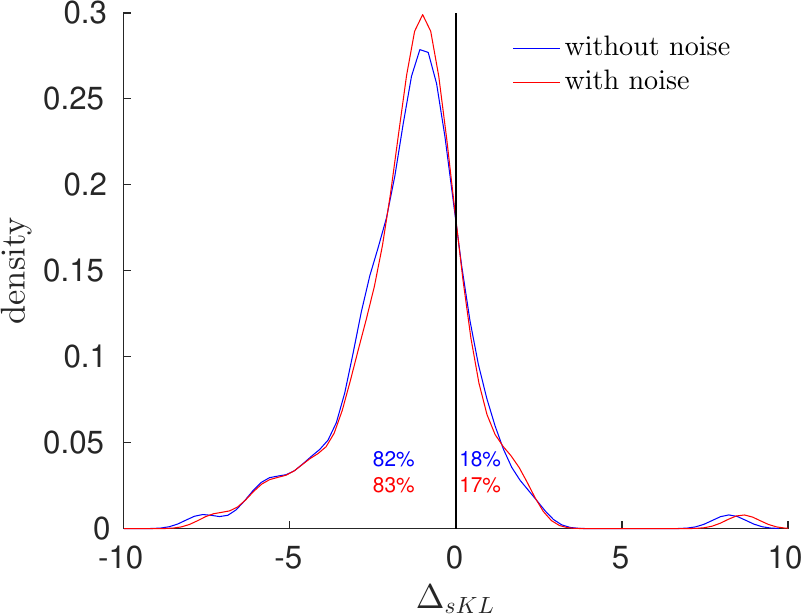}
	\caption{ARCH(1): Estimated density of the difference $\DeltaKL$
		between the symmetrised Kullback-Leibler divergence for linear
		LFIRE and synthetic likelihood with $n_{\theta} = n_{m} = 1000$,
		averaged over $100$ simulated data sets. A negative value of
		$\DeltaKL$ indicates that the proposed LFIRE method has a smaller
		divergence and thus is performing better. Depending on whether
		irrelevant summary statistics are absent (blue) or present (red)
		in the proposed method, it performs better than synthetic
		likelihood for 82\% or 83\% of the simulations. These results
		correspond to p-values from a Wilcoxon signed-rank test for
		pairwise median comparison of $1.06 \cdot 10^{-11}$ and $1.42
		\cdot 10^{-11}$, respectively. The densities were estimated with
		a Gaussian kernel density estimator with bandwidth 0.5.}
	\label{fig:arch_case_b_case}
\end{figure}

\begin{figure}
	\centering
	\includegraphics[width = 0.7\textwidth]{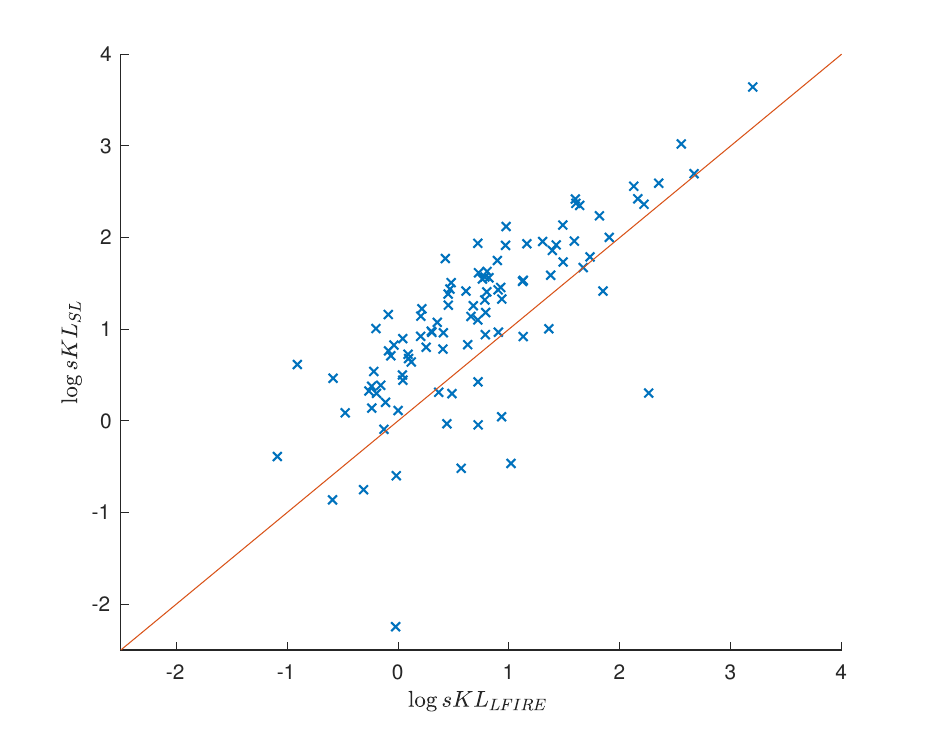}
	\caption{ARCH(1): A scatter plot of the logarithm of the
		symmetrised Kullback-Leibler divergence (sKL) of the
		proposed method and synthetic likelihood for $n_{\theta} =
		n_{m} = 1000$, evaluated over $100$ simulated data sets. The
		red line represents hypothetical equal performance of the
		two methods: we see that a substantial majority of
		simulations fall above this line, indicating better
		performance of the LFIRE method.}
	\label{fig:arch_scatter}
\end{figure}

\begin{figure}
	\centering
	\subfloat[$\log r$]{\includegraphics[width=0.475\textwidth]{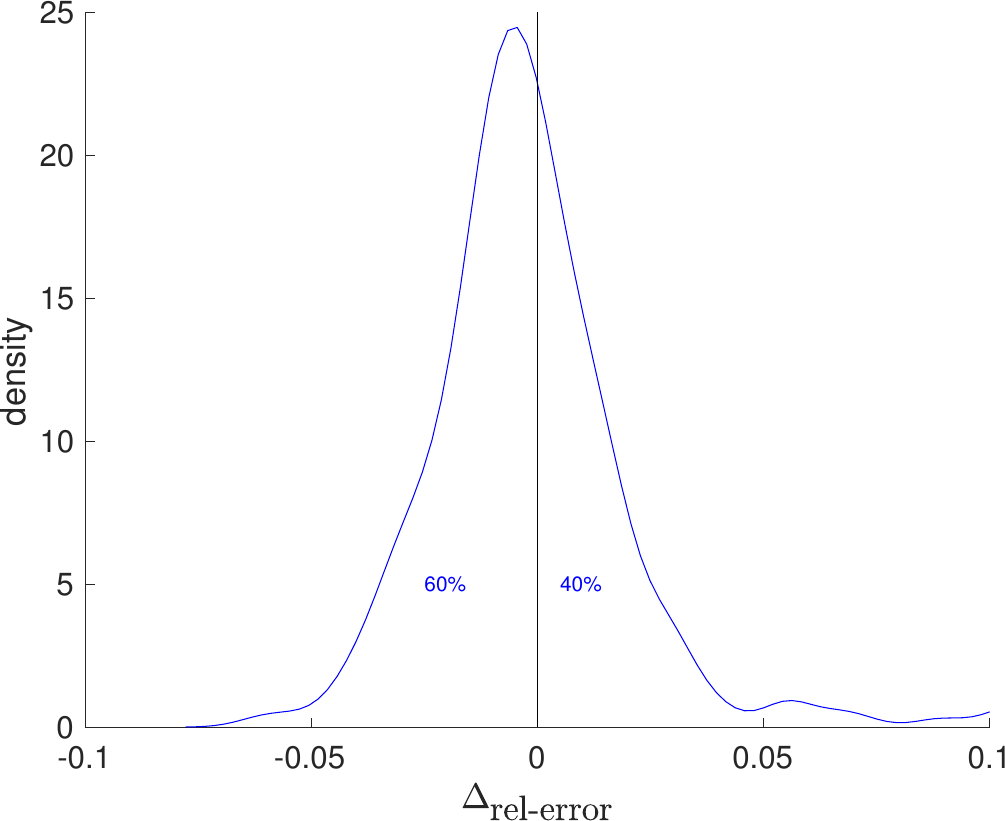}
		\label{fig:Ricker_logr_rel_err}}\\
	\subfloat[$\sigma$]{\includegraphics[width=0.475\textwidth]{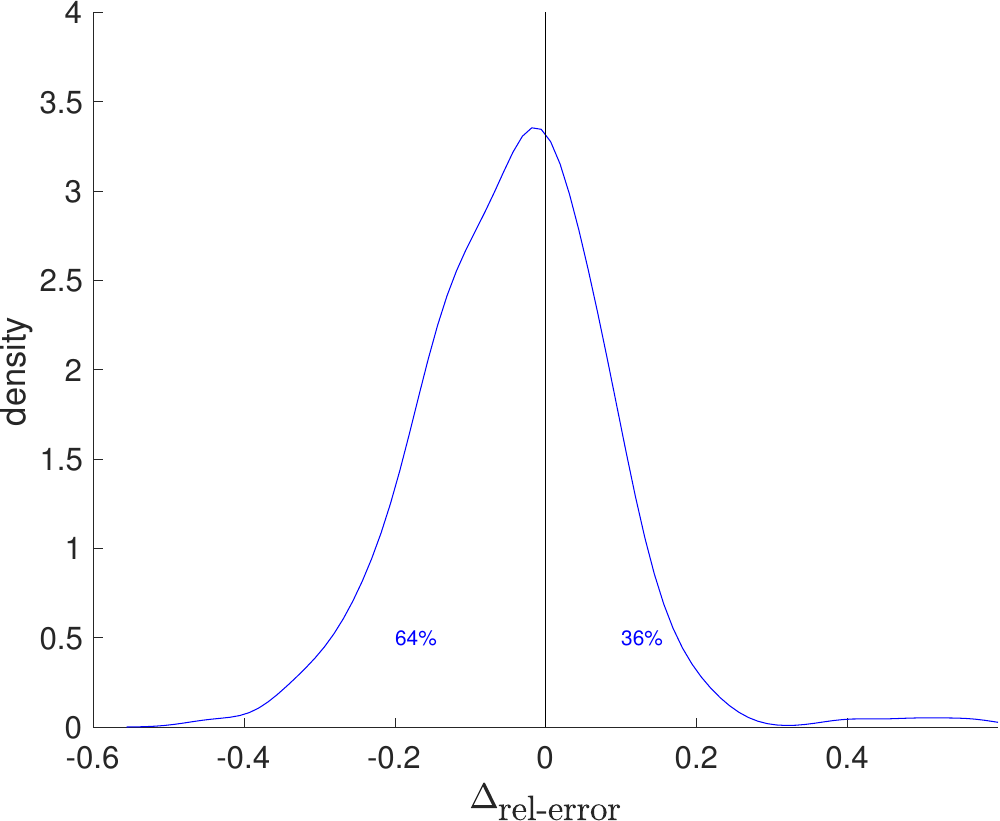}\label{fig:Ricker_sigma_rel_err}}
	\hfill
	\subfloat[$\phi$]{\includegraphics[width=0.475\textwidth]{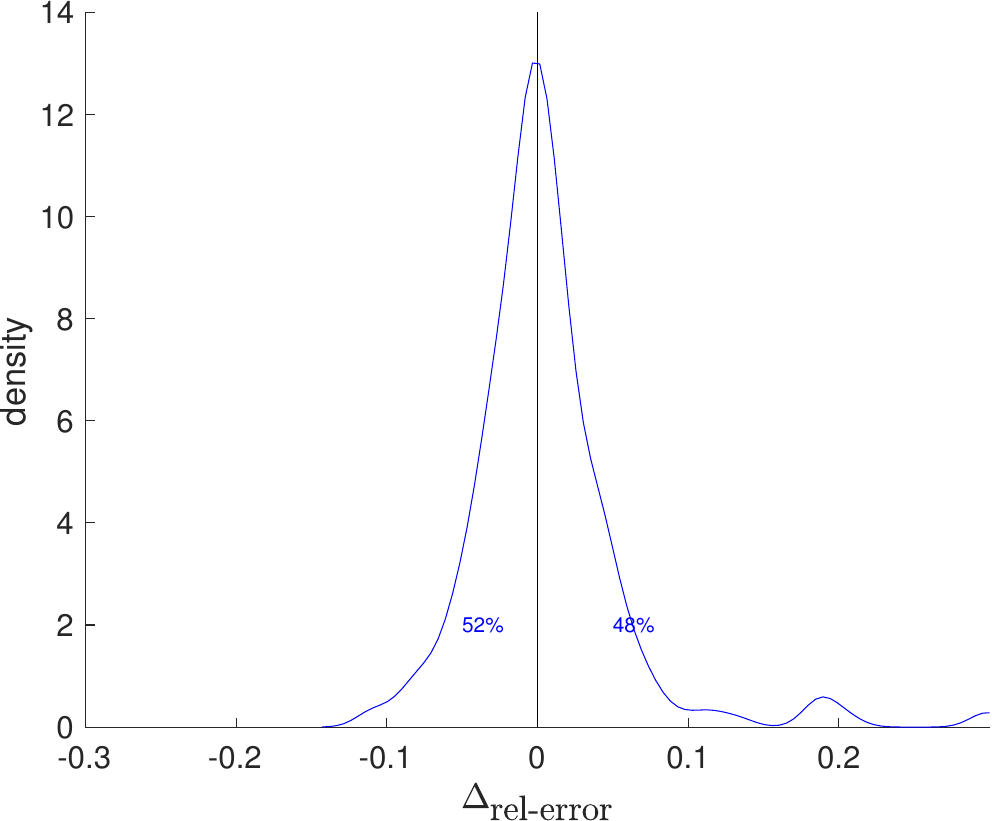}\label{fig:Ricker_lambda_rel_err}}
	\caption{Ricker model: Empirical pdf of $\Deltarelerr$ for the
		posterior mean of the parameters (a) $\log r$, (b) $\sigma$ and (c)
		$\phi$, compared against the mean from a rejection ABC algorithm,
		which drew $10,000$ samples at an acceptance rate of 0.02. More area
		under the curve on the negative side of the x-axis indicates a
		better performance of the proposed method compared to the synthetic
		likelihood. We used linear LFIRE in Algorithm \ref{algo:approx_posterior} and
		synthetic likelihood with $\ntheta=n_m = 100$ to estimate the
		posterior pdf for $250$ simulated observed data sets. The densities
		in (a--c) were estimated using a Gaussian kernel density estimator
		with bandwidth 0.01, 0.07 and 0.02, respectively. Using a
		nonparametric Wilcoxon signed-rank test for pairwise median
		comparison, these results correspond to p-values of $ 0.0074$, $4.99
		\cdot 10^{-8}$ and $0.7748$, respectively. The plots thus show that
		linear LFIRE is more accurate than synthetic likelihood in estimating the
		posterior mean of $\log r$ and $\sigma$ while the performance is
		similar for $\phi$.  }
	\label{fig:Ricker_rel_error}
\end{figure}

\begin{figure}
	\centering
	\subfloat[$\theta_1$]{\includegraphics[width=0.5\textwidth]{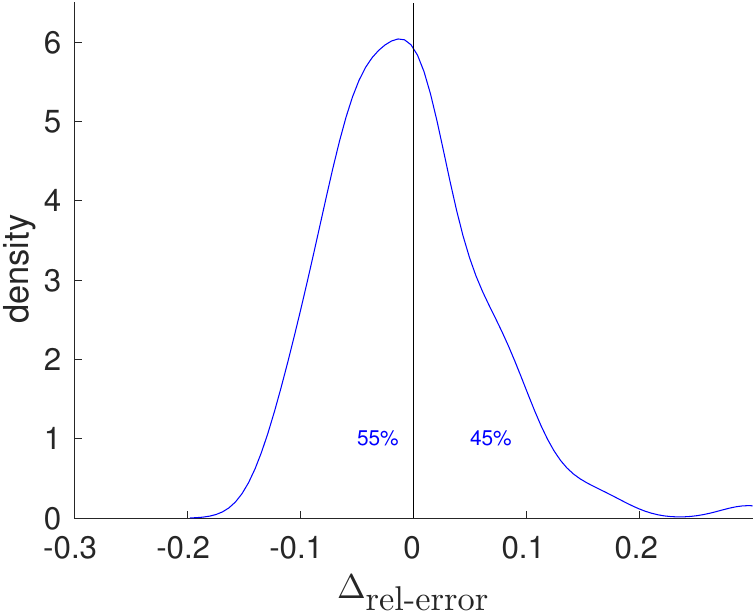}\label{fig:Lorenz_theta1_rel_err}}
	\hfill
	\subfloat[$\theta_2$]{\includegraphics[width=0.5\textwidth]{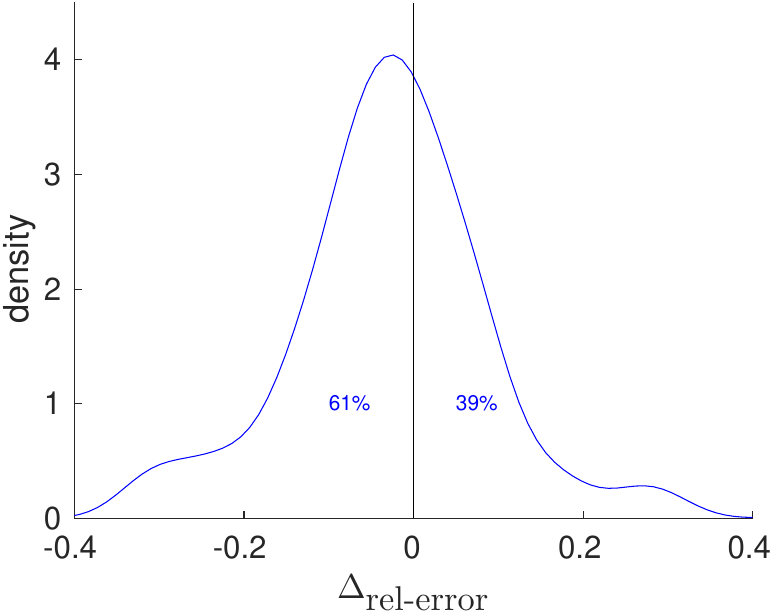}\label{fig:Lorenz_theta2_rel_err}}
	\caption{Lorenz model: Empirical pdf of $\Deltarelerr$ for the
		posterior mean of the parameters (a) $\theta_1$ and (b)
		$\theta_2$, compared against the mean from a rejection ABC
		algorithm, which drew $48,000$ samples at an acceptance rate of
		0.016. More area under the curve on the negative side of the
		x-axis indicates a better performance of the proposed method. We
		used linear LFIRE in Algorithm \ref{algo:approx_posterior} and synthetic
		likelihood with $\ntheta=n_m = 100$ to estimate the posterior pdf
		for 250 simulated observed data sets. The densities in (a--b) were
		estimated using a Gaussian kernel density estimator with bandwidth
		0.025 and 0.037, respectively. Using a nonparametric Wilcoxon
		signed-rank test for pairwise median comparison, these results
		correspond to p-values of $4.59 \cdot 10^{-9}$ and $6.92 \cdot
		10^{-25}$, respectively. The plots thus show that linear LFIRE is more
		accurate than synthetic likelihood in estimating the posterior
		mean of the parameters of the Lorenz model.  }
	\label{fig:Lorenz_rel_error}
\end{figure}

\begin{figure}
	\centering
	\subfloat[$\log r$]{\includegraphics[width=0.475\textwidth]{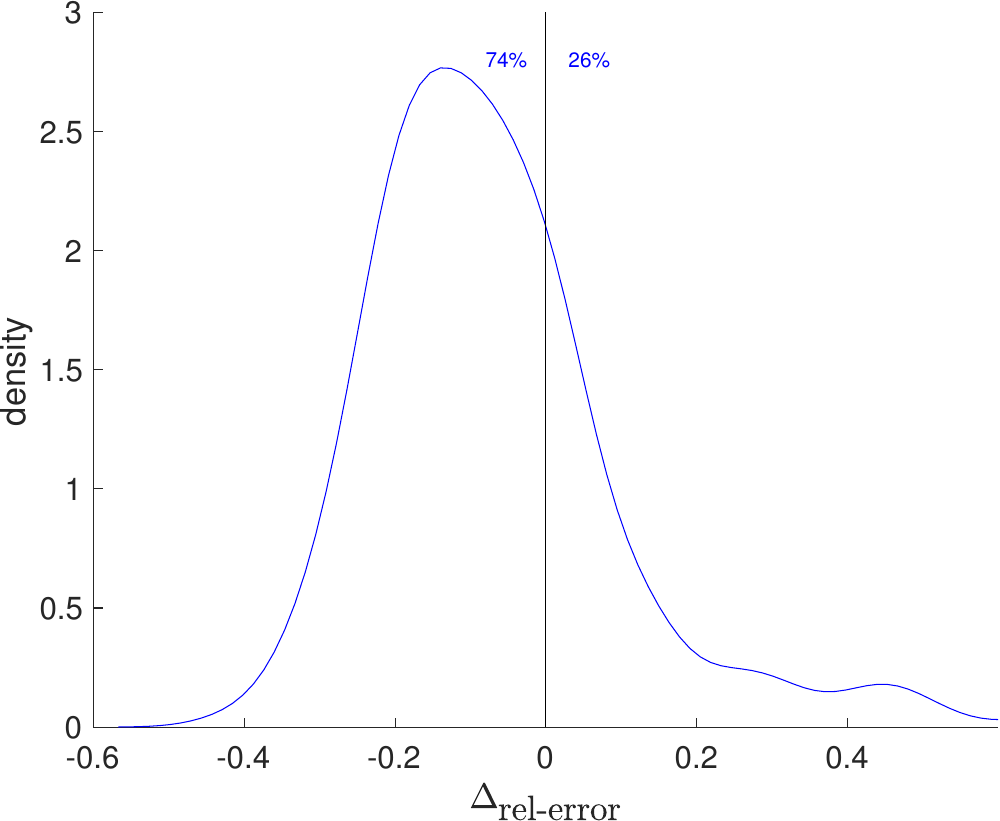}}\\
	\subfloat[$\sigma$]{\includegraphics[width=0.475\textwidth]{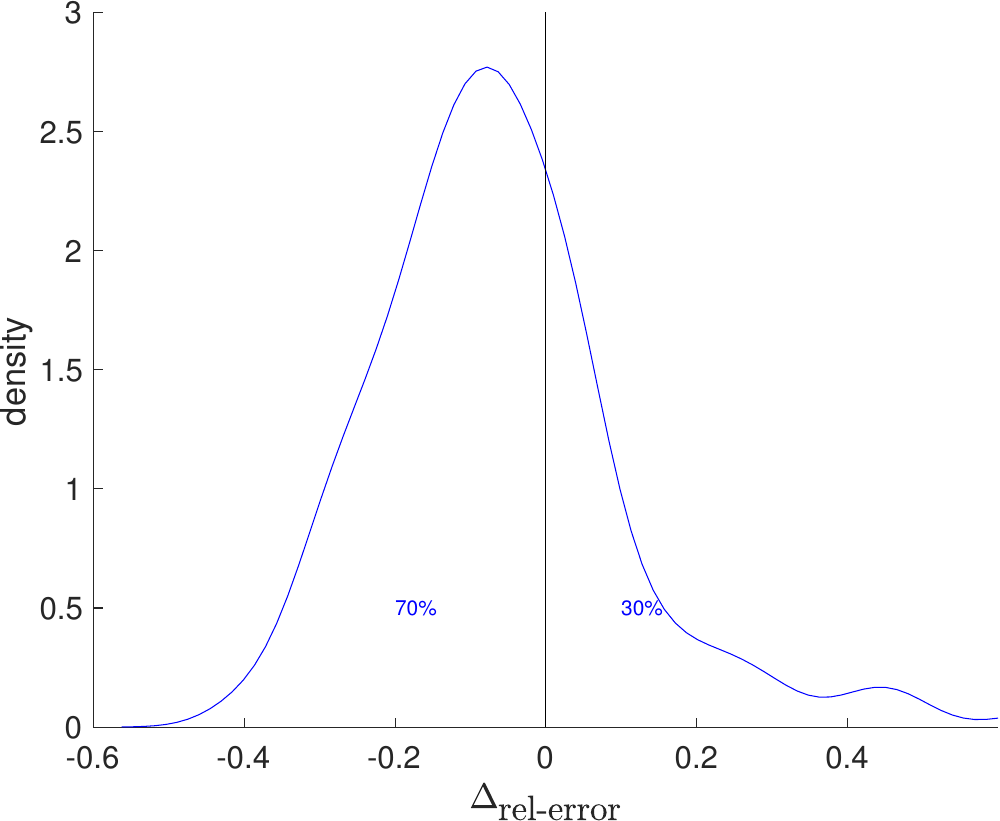}}
	\hfill
	\subfloat[$\phi$]{\includegraphics[width=0.475\textwidth]{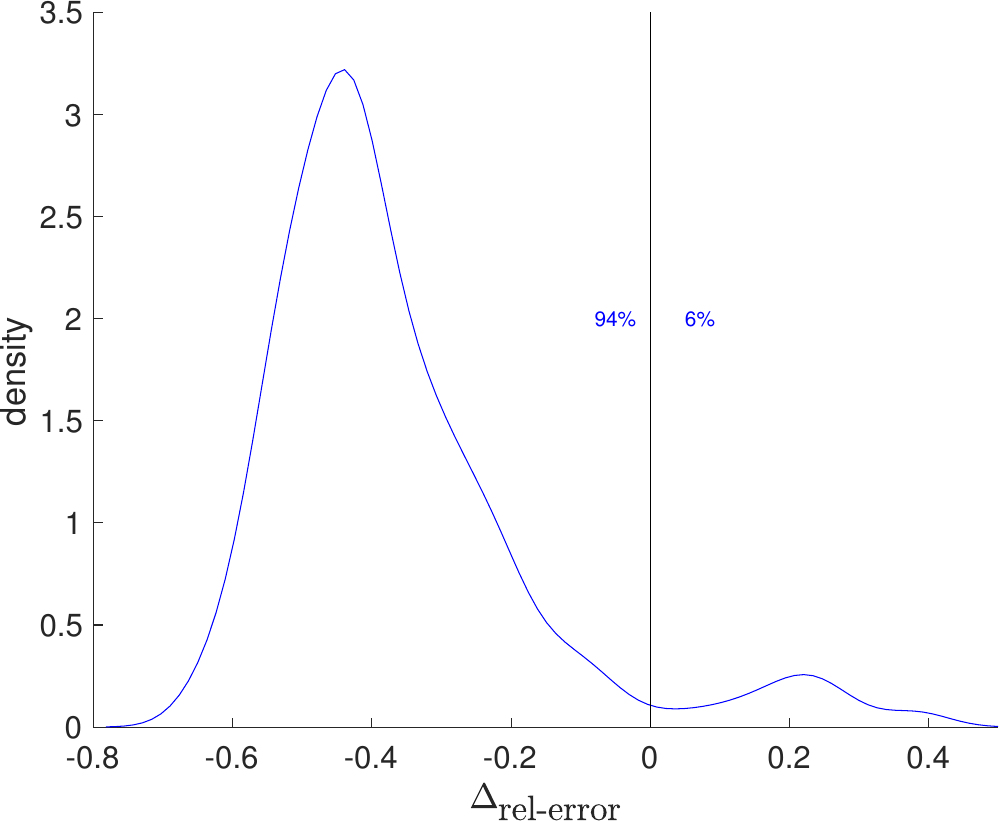}}
	\caption{\label{fig:Ricker_rel_error_std} Ricker model: Empirical pdf
		of $\Deltarelerr$ for the posterior standard deviations of the
		parameters (a) $\log r$, (b) $\sigma$ and (c) $\phi$.  Setup is as
		in Figure \ref{fig:Ricker_rel_error}. Using a nonparametric Wilcoxon
		signed-rank test for pairwise median comparison, these results
		correspond to p-values of $2.48 \cdot 10^{-14}, 1.76 \cdot 10^{-11},
		5.98 \cdot 10^{-40}$, respectively. The plots thus show that linear LFIRE
		is more accurate than synthetic likelihood in estimating the
		posterior standard deviation (uncertainty) of the parameters of the Ricker model.
	}
\end{figure}

\begin{figure}
	\centering
	\subfloat[$\theta_1$]{\includegraphics[width=0.475\textwidth]{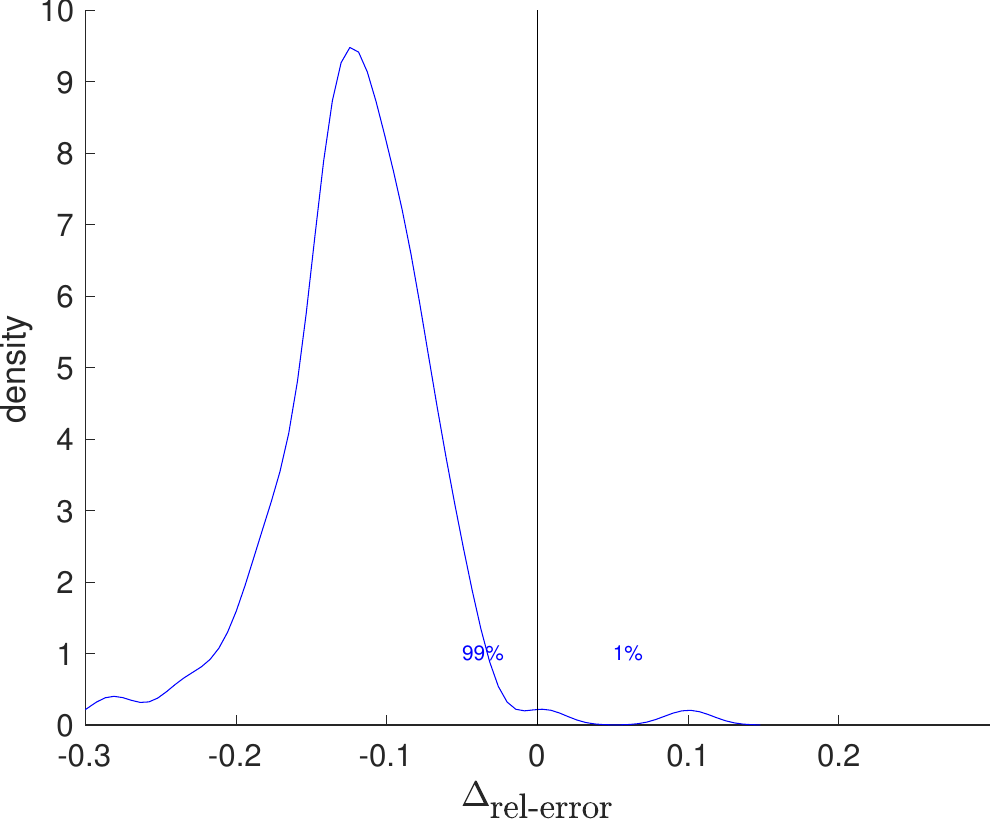}}
	\hfill
	\subfloat[$\theta_2$]{\includegraphics[width=0.475\textwidth]{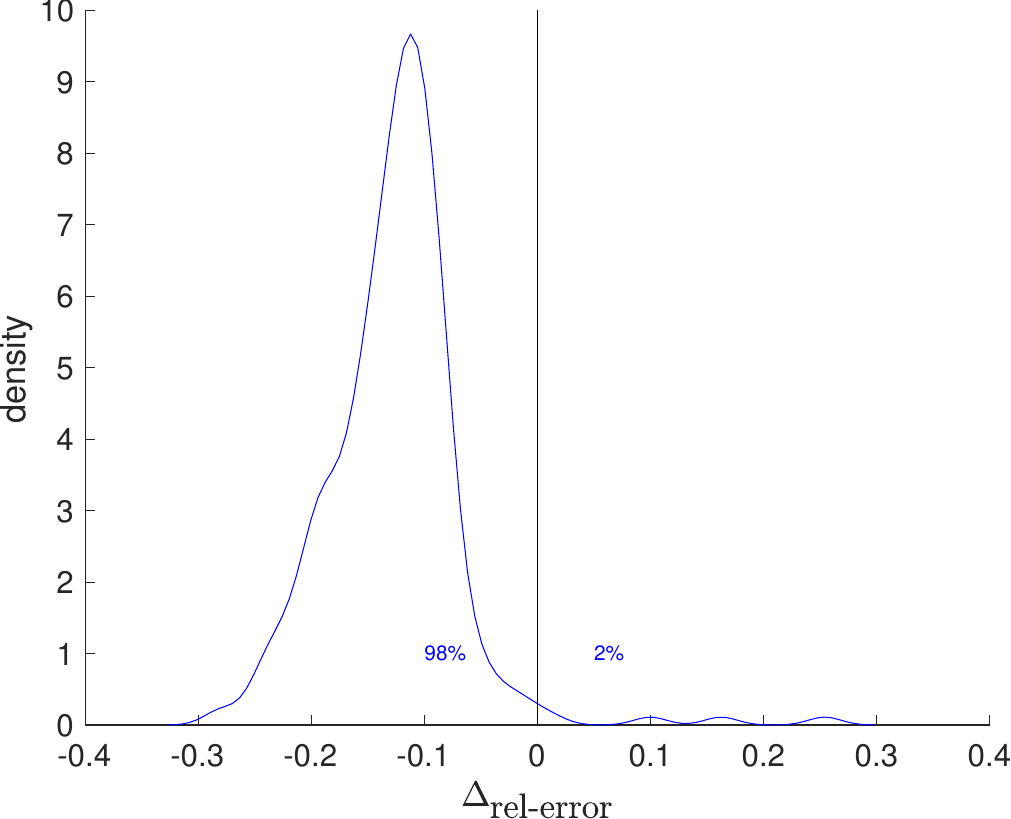}}
	\caption{\label{fig:Lorenz_rel_error_std} Lorenz model: Empirical
		pdf of $\Deltarelerr$ for the posterior standard deviations of the
		parameters (a) $\theta_1$ and (b) $\theta_2$. Setup is as in
		Figure \ref{fig:Lorenz_rel_error}. Using a nonparametric Wilcoxon
		signed-rank test for pairwise median comparison, these results
		correspond to p-values of $6.29 \cdot 10^{-42}$ and $3.8 \cdot
		10^{-40}$, respectively. The plots thus show that linear LFIRE is more
		accurate than synthetic likelihood in estimating the posterior
		standard deviation (uncertainty) of the parameters of the Lorenz
		model.}
\end{figure}

\begin{figure}
	\centering \includegraphics[width = 0.5\textwidth]{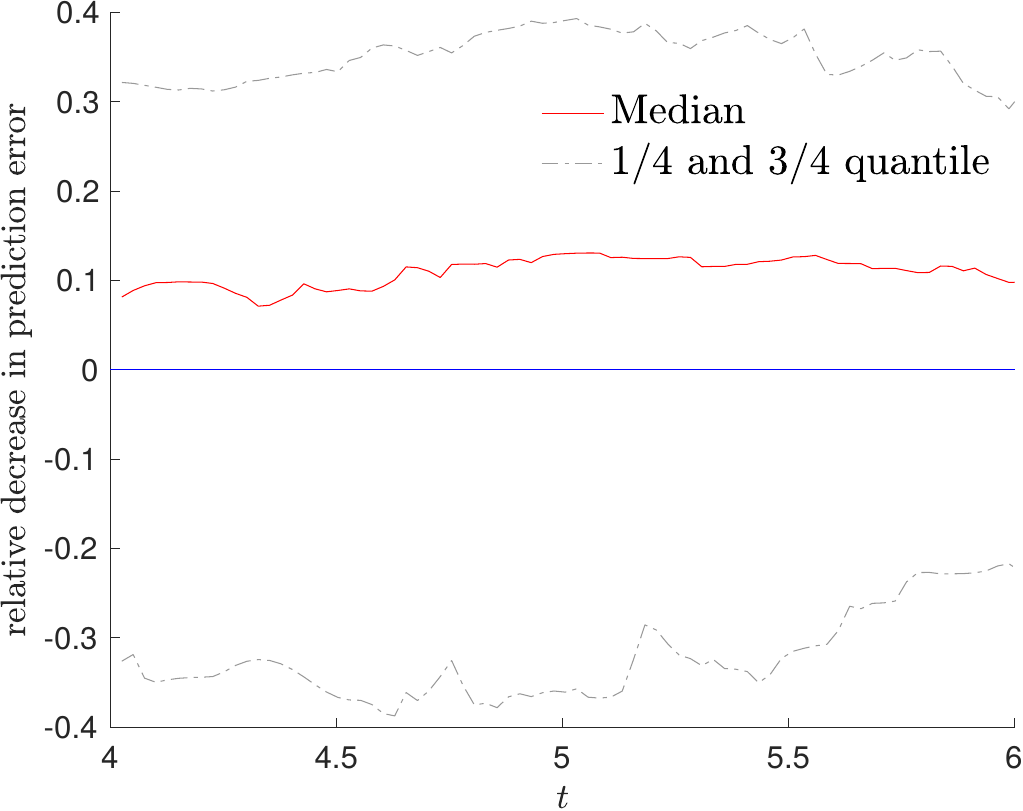}
	\caption{Lorenz Model: Median, $1/4$ and $3/4$ quantile
		of the relative decrease in the prediction error
		$\zeta^{(t)}$ for $t \in [4,6]$ corresponding to 1 to 10 days in
		the future. We used linear LFIRE in Algorithm \ref{algo:approx_posterior} and
		synthetic likelihood with $\ntheta=n_m = 100$ to estimate the
		posterior pdf. As the median is always positive, the proposed
		method obtains, on average, a smaller prediction error than
		synthetic likelihood.}
	\label{fig:lorenz_predic_err}
\end{figure}

\fi

\bibliographystyle{ba}
\bibliography{reference}

\begin{acknowledgement}
The work was partially done when OT, RD and MUG were at the Department
of Biostatistics, University of Oslo, Department of Computer Science,
Aalto University, and the Department of Mathematics and Statistics,
University of Helsinki, respectively. The work was financially
supported by the Academy of Finland (grants 294238 and 292334, and the
Finnish Centre of Excellence in Computational Inference Research
COIN). The authors thank Chris Williams for helpful comments on an
earlier version of the paper and gratefully acknowledge the
computational resources provided by the Aalto Science-IT project. RD
was supported by Swiss National Science Foundation grant
no.\ $105218\_163196.$ JC and OT were supported by ERC grant no.\ 742158.

\end{acknowledgement}

\end{document}